\documentclass[10pt,twocolumn,letterpaper]{article}

\usepackage[pagenumbers]{cvpr} 

\usepackage{graphicx}
\usepackage{booktabs}
\usepackage{times}
\usepackage{epsfig}
\usepackage{graphicx}
\usepackage{array}
\usepackage{amsmath}
\usepackage{amssymb}
\usepackage{tabularx}
\usepackage{multirow}
\usepackage{booktabs}
\usepackage{arydshln}
\usepackage[dvipsnames]{xcolor}
\usepackage{fontawesome}
\usepackage{caption}
\usepackage{algorithm2e}
\usepackage{enumitem}
\usepackage{pifont}
\usepackage{xspace}
\usepackage{tabularx}
\usepackage{multirow}
\usepackage{caption}
\usepackage{booktabs}
\usepackage{cite}
\usepackage{bm}
\usepackage{comment}

\definecolor{cvprblue}{rgb}{0.21,0.49,0.74}
\usepackage[pagebackref,breaklinks,colorlinks,citecolor=cvprblue]{hyperref}

\DeclareMathOperator*{\argmin}{arg\,min}
\definecolor{ourgreen}{RGB}{56,87, 35}
\definecolor{ourorange}{RGB}{184,97, 35}

\def\paperID{346} 
\def\confName{3DV\xspace}
\def\confYear{2025\xspace}

\title{Maps from Motion (MfM): Generating 2D Semantic Maps from Sparse Multi-view Images} 

\author{
Matteo Toso$^1$ Stefano Fiorini$^1$ {Stuart James}$^{1,2}$ Alessio Del Bue$^1$ \\
$^1$Istituto Italiano di Tecnologia (IIT) \\
$^6$Durham University\\
{\tt\small matteo.toso@iit.it}
}

\begin{document}

\makeatletter
\let\@oldmaketitle\@maketitle
\renewcommand{\@maketitle}{\@oldmaketitle
  \centering\includegraphics[width=0.8\linewidth, trim={1cm 4.5cm 0cm 0cm}, clip] {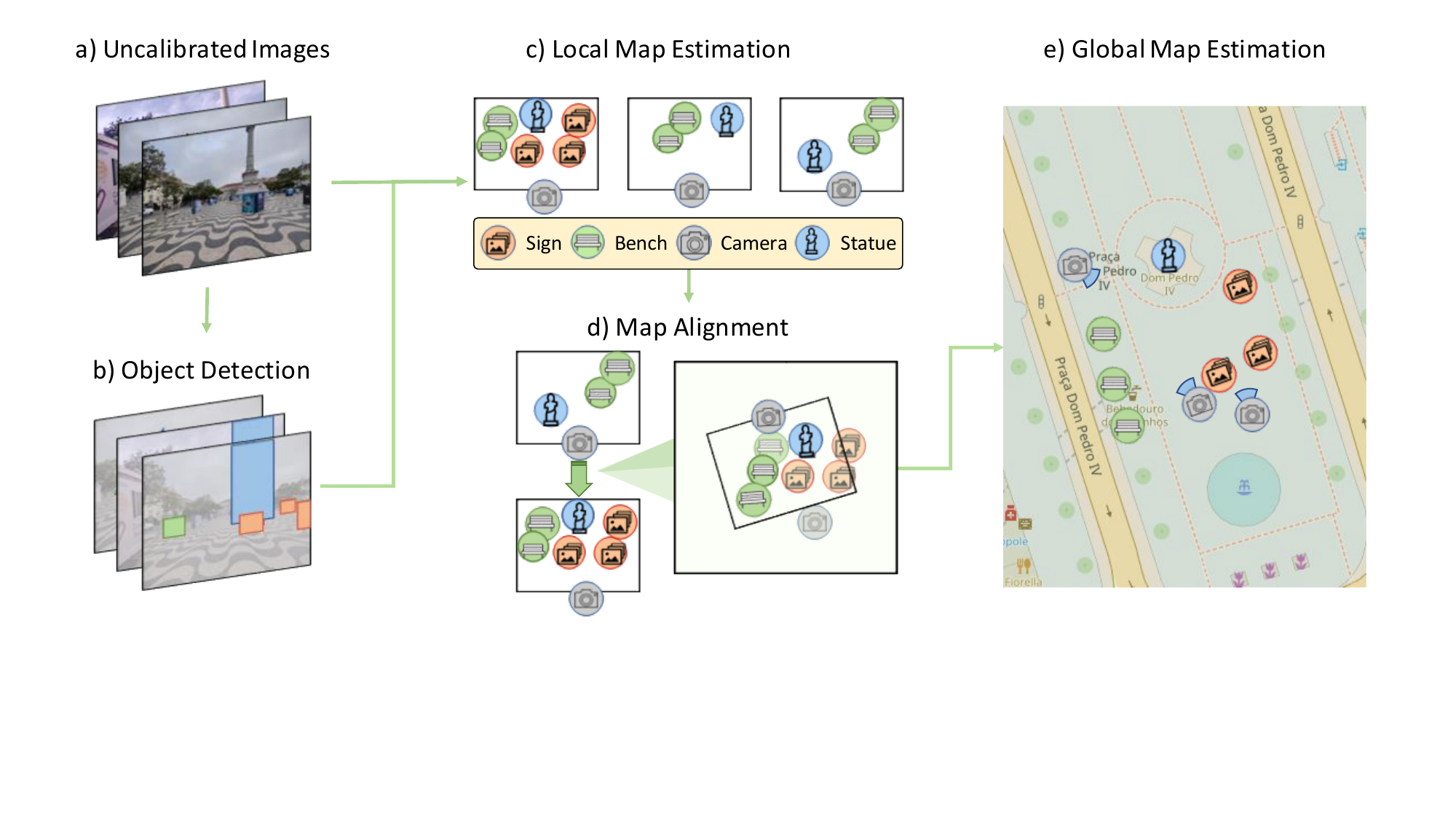}
    \captionof{figure}{\emph{The Maps from Motion (MfM) Concept}.  From a set of uncalibrated images (a) we extract 2D detections of static urban objects (b), and generate local top-down 2D maps representing the spatial arrangement of the objects with respect to each image (c). We then learn how to register all local maps in the same reference frame (d), to generate a common global map with all objects present in the scene (e).}
    \label{fig:concept}\bigskip}
\makeatother

\maketitle

\begin{abstract}
World-wide detailed 2D maps require enormous collective efforts. \textit{OpenStreetMap} is the result of 11 million registered users manually annotating the GPS location of over 1.75 billion entries, including distinctive landmarks and common urban objects. At the same time, manual annotations can include errors and are slow to update, limiting the map's accuracy. \textit{Maps from Motion} (MfM) is a step forward to automatize such time-consuming map making procedure by computing 2D maps of semantic objects directly from a collection of uncalibrated multi-view images. From each image, we extract a set of object detections, and estimate their spatial arrangement in a top-down local map centered in the reference frame of the camera that captured the image.
Aligning these local maps is not a trivial problem, since they provide incomplete, noisy fragments of the scene, and matching detections across them is unreliable because of the presence of repeated pattern and the limited appearance variability of urban objects. We address this with a novel graph-based framework, that encodes the spatial and semantic distribution of the objects detected in each image, and learns how to combine them to predict the objects' poses in a global reference system, while taking into account all possible detection matches and preserving the topology observed in each image. 
Despite the complexity of the problem, our best model achieves global 2D registration with an average accuracy within $4$ meters (\ie below GPS accuracy) even on sparse sequences with strong viewpoint change, on which COLMAP has an $80\%$ failure rate. 
We provide extensive evaluation on synthetic and real-world data, showing how the method obtains a solution even in scenarios where standard optimization techniques fail.
\end{abstract}

\section{Introduction}
\label{sec:intro}

2D semantic maps provide top-down abstract representations of an environment annotated with the location of easily identifiable landmarks, and play an important role in everyday life. 
In most public spaces, like museums and parks, we are used to finding our way via minimalist maps that mark our current location as a red, ``You are here \textcolor{red}{\faMapMarker}'' dot.

In recent years, works like OrienterNet~\cite{sarlin2023orienternet}, SNAP~\cite{sarlin2023snap} and Flatlandia~\cite{toso2023you} have highlighted the advantages of using semantic maps in Computer Vision. They are more storage efficient than traditional 3D maps from Lidar or photogrammetry~\cite{romeinaday, moulon2016openmvg, Schonberger_2016_CVPR}, requiring $1-10\%$ of the memory used by a set of reference images or a 3D point cloud~\cite{toso2023you}. Moreover, semantic maps provide an abstract representation that is robust to temporal changes.  
While a 2D map might sound limiting, in several practical scenarios like autonomous cars or robotics, the cameras have roll angle $0$ and y-axis colinear with the gravity direction~\cite{Toft_2018_ECCV}, and the height of the cameras from the ground is constant~\cite{1638022}. In such cases, localization as GPS location and a viewing direction is sufficient.
Moreover, previous work has shown that bird-eye-view (BEV) maps provide enough information to achieve accurate localization~\cite{sarlin2023orienternet}. We explore the possibility of annotating global 2D maps using an even more abstract representation, composed only of the spatial top-down view layout of the objects observed in the images.

Existing approaches for generating 2D maps with object annotations present several limitations, in terms of time (\eg requiring manual user annotations~\cite{osm}), computational cost (\eg large SfM reconstructions ~\cite{toso2023you}), and specialized additional data (\eg aerial images ~\cite{sarlin2023snap}).
In contrast, we suggest that directly reasoning in 2D to combine partial maps from different viewpoints is a more efficient approach.

Given a sparse set of images, we frame the reconstruction of the 2D map as the registration of partial maps, based on the estimated arrangement of the objects detected in each image.
As shown in Figure~\ref{fig:concept}, we take a set of uncalibrated, unsorted images of the scene (\ref{fig:concept}.a) and extract from them detections of common urban objects (\ref{fig:concept}.b). Then, using the nominal camera intrinsics and monocular depth estimation, we  generate local maps representative of the observed objects' layout in the camera reference system (\ref{fig:concept}.c). We want to align these local maps in a common reference system (\ref{fig:concept}.d), resulting in a global map (\ref{fig:concept}.e); this, however, requires inferring a transformation (roto-translation and scale) from each map to a common global reference frame. We name this novel problem \textit{Maps from Motion} (MfM). 

The proposed task presents several difficulties: using only the estimated location of objects is an efficient representation~\cite{toso2023you}, but it carries less information than traditional BEV images. 
Representing objects as a single coordinates also limits the semantic classes available, since larger elements like roads and buildings, while typically used in semantic maps, are too large to be localized as a single point without making the problem unstable.
At the same time, using sparse images and focusing common urban objects, that typically have plain and standardized appearance, makes establishing object matches from the input images unreliable. Given the success of Graph Neural Network (GNN) to address geometrical reasoning problems~\cite{fuchs2020se, giuliari2023positional, scarpellini2024diffassemble}, we frame MfM as a graph problem, assigning a node to each detection and attempting to regress its location in the global map. In this formulation, we use same-map edges to force the network to preserve the topology of each local map, and same-class edges to account for all possible detections matches, instead of explicitly matching the input detections. This amounts to training a network to find the best alignment between the local maps, while preserving the object's layout observed in each image. To investigate the ability of graph networks to solve the MfM problem, we compare several architectures, with and without an attention mechanism.
Through experiments on the Flatlandia dataset~\cite{toso2023you}, 
we show that, 
despite the noisy detections and the absence of explicit cross-image detection matches, can achieve object and camera localization accuracy comparable to COLMAP, while achieving a $60\%$ lower failure rate on sparse sequences with strong viewpoint changes. Even in this challenging scenario, the best-performing implementation of our solution achieves a median localization error of less than $4$ meters, better than standard GPS accuracy ($4.9$~m\footnote{\href{https://www.gps.gov/systems/gps/performance/accuracy/}{www.gps.gov/performance/accuracy}}). Our contributions are the following:
\begin{itemize}
    \item We introduce a new problem (MfM) that provides an object scene map from a sparse set of uncalibrated images to automatize 2D map making procedures.
    \item We propose a new graph structure to address the MfM correspondence and registration problem, along with a GNN framework that estimates the positions of cameras and objects in a 2D global reference frame.
    \item We provide a new dataset and an evaluation protocol for MfM 
    demonstrating the feasibility of the problem, and offer a comparison against  
    relevant baselines.
\end{itemize}

\section{Related Work} 
\label{sec:relatedwork}
We discuss 
the creation (Section~\ref{sec:relatedwork:map_create}), representation (Section~\ref{sec:relatedwork:map_represent}), and use (Section~\ref{sec:relatedwork:map_use}) of 2D semantic maps.

\subsection{Creation of Semantic Maps}\label{sec:relatedwork:map_create}
The creation of semantic maps has often been intertwined with 3D modeling, as their models - \eg 3D point-cloud reconstructions from Structure from Motion (SfM) - provide enough information to allow localizing 3D objects~\cite{7919240, Li_2021_ICCV,toso2023you}. 
Alternatively, 3D objects can be parameterized as ellipsoids~\cite{zinshal02975379}, which, in turn, can also be used jointly for camera pose estimation from elliptical detections~\cite{crocco2016structure, 8440105, Gay_2017_ICCV}. 
This parameterization is also used in
SLAM~\cite{Nicholson_2018_CVPR_Workshops, 8353862, 8460816}. 
Such methods result in accurate models, but they generally are computation and memory intensive, and prone to failure if the images are too sparse or have large viewpoint changes.

Other methods use object detections to create object-maps. If camera calibration is available, an option is to use 3D projection 
\cite{nassar2019simultaneous} and refining using graph neural networks \cite{nassar2020geograph} to produce an object-level map. However, these approaches impose strong limitations on the diversity of objects (i.e. trees) and the camera parameters they can handle. Object detection are also 
used to re-identify buildings for camera localization~\cite{Xue_2022_CVPR}, matching detections across images~\cite{cathrin}, and making feature matching more robust~\cite{benbihi2022}. 

Alternatively, some works have focused on on generating accurate bird-eye-view maps from input images~\cite{Saha2022CVPR, electronics12245017} or on improving maps' accuracy by fusing street-level images with additional sensor (\eg like satellite or aerial images~\cite{sarlin2023snap}).  Such additional data provide helpful localization cues, but aerial data and fleets of sensorized cars are viable only for large enterprises, limiting accessibility. 

Finally, platforms like OpenStreetMaps~\cite{osm} generate user-annotated semantic maps via crowd-sourcing. This provides large maps, but it is time and resource expensive, the annotation density varies significantly from area to area, and conflicting or old annotations can result in outdated maps.
Unlike these methods, we directly compute minimalist 2D semantic maps from images without an intermediate 3D model, characterizing the local maps as sets of 2D coordinates with a class label. This allows investigating the solubility of MfM from minimal information.

\subsection{Graph Representation of scenes}\label{sec:relatedwork:map_represent}
The use of graphs to represent scenes is most common
in scene graphs from image~\cite{gu2019scene,garg2021unconditional,graph2scene2021} or 3D models~\cite{armeni20193d,gay2019visual}. Here, the content of a scene is encoded in graph form and passed to a GNN, trained on tasks like relational labeling \cite{gay2019visual}, object localization~\cite{giuliari2022spatial} and robot navigation~\cite{Ravichandran2022Scenenavigation}.

Moreover, many works use graph to encode the spatial representation of 2D Map data 
for representing and inferring map attributes~\cite{he2020roadtagger,bandil2021geodart,yin2020multi,huang2022msen} or for downstream tasks like traffic management~\cite{qin2020graph}.
Other works, closer to MfM, align a 3D representations to a 2D reference map~\cite{chen2024integrating,li2023unleashing}, propagating the 2D information to elements of the 3D model. This problem is simpler than MfM, as it involves a single alignment problem instead of multiple ones (each local map to a common reference frame). 

GNNs are also used to solve the 3D camera pose estimation problems. Recent works have formalized this problem as motion averaging~\cite{purkait2020neurora, yew2020-RobustSync, posernet_eccv2022} and Bundle Adjustment (BA)~\cite{Brynte2023LearningSW} through Graph Neural Networks (GNNs)~\cite{kipf2016semi, liu2020graphsage, fiorini2023sigmanet, fiorini2024graph}, typically modeling the scene as a graph of connected cameras to update or regress the 6DoF camera pose.

These approaches rely on 3D data to make their problems more treatable. Such data, however, is often not available, making it appealing to find a solution to the challenging problem of sparse scene estimation problem in 2D.

\subsection{Use of Semantic Maps}\label{sec:relatedwork:map_use}
Most works exploiting semantic maps are used for visual localization in 3DoF, \ie as a 2D coordinate and a viewing angle. This pose conveys less information than a camera pose in 3D, but for some applications (\eg in the automotive sector~\cite{Shi_2023_ICCV}) it is enough. Recent works~\cite{9635972,noe2020eccv,Vojir_2020_ACCV, sarlin2023orienternet} use reference maps from OpenStreetMaps~\cite{osm}, which provide vector tiles with objects (polygonal areas, multi-segment lines, or single points) representing various urban elements (\eg building outlines, roads, and objects, respectively). Localization is then performed by comparing embeddings of the visual query and of the reference information extracted at different map locations~\cite{9635972, noe2020eccv, Vojir_2020_ACCV} or regresses the 3DoF pose from a visual query by comparing bird's-eye view neural maps generated via GNNs from a query~\cite{sarlin2023orienternet}.

\section{Methodology}\label{sec:method}

\begin{figure*}[t]
        \centering
        \includegraphics[width=0.90\linewidth, trim={0 3.5cm 0cm 0cm}, clip] {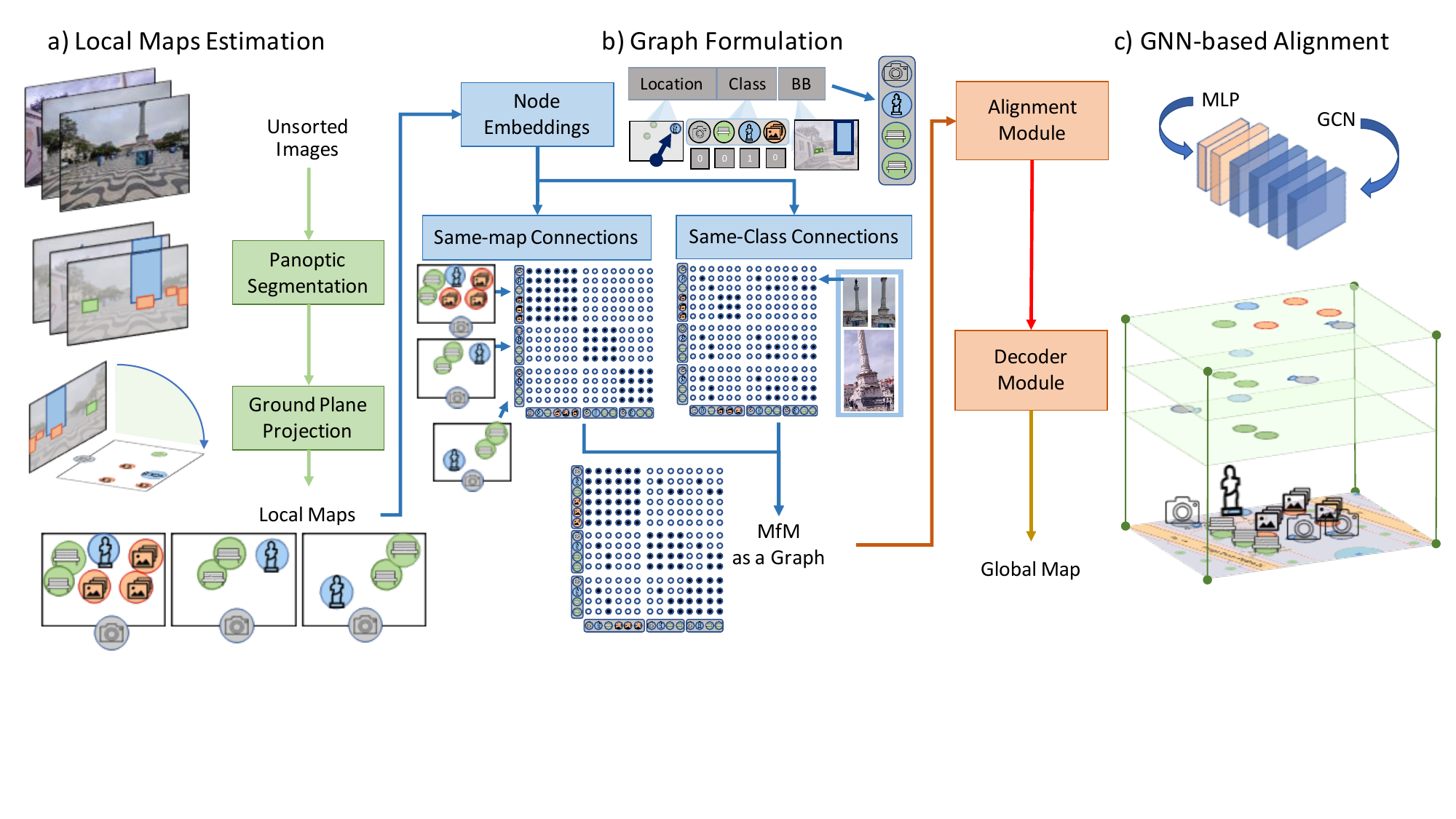}
    \caption{\emph{The MfM Pipeline}. a) We extract 2D maps representing the spatial arrangement of detected objects, from the image's point of view. b) The maps are encoded as a graph, with a node for each detection, and edges connecting detections from the same image (Same-Map) or with the same class label (Same-class). c) A GNN predicts the location of all object and the cameras in one reference frame.}
    \label{fig:themethod}
\end{figure*}


MfM addresses the problem of generating a \textbf{2D global map} from a set of sparse and wide-baseline images. As illustrated in Figure~\ref{fig:themethod}, this is done across three main steps. First, the images are turned into \textbf{local maps} (Figure~\ref{fig:themethod}.a), representing the spatial arrangement of the objects detected in the images  and their classes. These maps are obtained by reprojecting the center of the object detections in 3D and then projecting it on the ground plane, as detailed in Section~\ref{sec:method.problem_setup}. MfM combines the local maps by forming a graph (Figure~\ref{fig:themethod}.b), in which the detections are represented as nodes, the local maps as subgraphs, and the cross-map connections as edges (Section~\ref{sec:method.graph_representation}). We then train a GNN to aggregate the information and regress the location of each detected object in the reference system of a  global map (Figure~\ref{fig:themethod}.c), while preserving the spatial layout of each local map.

\subsection{Local Map Estimation}\label{sec:method.problem_setup}
Given a set of $K$ images $\mathcal{I}=\{I_i\}_{i\in[1,\dots,K]}$, we estimate a set of local maps $\mathcal{M} = \{M_i\}_{i\in[1,\dots,K]}$. First, we use the Panoptic object detection algorithm~\cite{Kirillov_2019_CVPR} to identify the objects in the scene. This results in a detection $o_{ij}$ with semantic label $l_{ij}$ for each object $j$ observed in the image. 
We then estimate the distance $d_{ij}$ of $j$ from image $i$ using a monocular depth estimation algorithm~\cite{ranftl2020towards} and averaging the predicted per-pixel depth over the detection, following ~\cite{toso2023you}. Finally, we represent the coordinates of the detection in the image as a bounding box $b_{ij}$. 
The use of monocular depth estimation and averaging the depth of the whole detection into a single point will introduce some noise, and for this reason we also evaluate the proposed approach using perfect depth (Sec.~\ref{sec:experiments:sfmo}) and perform ablation tests on the effect of noise on the local maps (Sec.~\ref{sec:experiments:ablations}). 

Finally, we generate the local maps by reprojecting the center of each bounding box in 3D, and then projecting it onto the scene's ground plane.
We then define the location $\mathbf{c}_{ij}\in \mathcal{R}^2$ of $j$ in the local map $M_i$ as:
\begin{equation}
    \begin{bmatrix} \mathbf{c}_{ij} \\ 1 \end{bmatrix} = \Pi \left(d_{ij} 
    \; K_i^{-1} \begin{bmatrix} \mathbf{b}_{ij} \\ 1 \end{bmatrix}\right),
\label{eq:point_to_detection}\end{equation}
where $\Pi$ is the projection from 3D to a horizontal plane orthogonal to the $I_i$, $\mathbf{b}_{ij}$ is the center of the bounding box $b_{ij}$, and $K_i$ is the intrinsic camera matrix associated with image $i$, obtained from the nominal characteristics of the camera. Equation~\eqref{eq:point_to_detection} projects the objects out of the image plane assuming a pinhole camera model, and we then define the 2D location of the camera that captured the image as the origin of the local map, \ie as coordinates $[0,0]$.
\subsection{Graph Formulation for the MfM Problem}\label{sec:method.graph_representation}

To solve MfM, we frame the alignment problem in graph form, addressing the alignment problem using the distribution of class labels and the spatial layout, instead of explicitly matching detections across views. 

First, we define for each image $I_i$ an undirected subgraph $G_i=\{V_i, E_i\}$, with nodes $V_i$ and edges $E_i$ . We assign a node to each object detection, and one to the camera associated with $I_i$. The edges $E_i$ connect all nodes $V_i$, making $G_i$ a fully connected graph. 
We then aggregate the local maps in a large, undirected graph $\tilde{\mathcal{G}} = \{ \tilde{V}, \tilde{E}\}$, whose nodes are defined as the union of all subgraph's nodes $\tilde{V} = \cup_{G_i} V_i$. The edges $\tilde{E}$ are defined as $\tilde{E} = (\cup_{G_i} E_i) \cup E_{l}$, where $E_l $ is the set of edges connecting all nodes associated to the same class label, \ie possible detection matches. 

From the set of edges $\tilde{E}$, we then define a binary adjacency matrix, denoted as $A\in \{0, 1\}^{|\tilde{V}|\times |\tilde{V}|}$; this is a block diagonal structure connecting object detections from the same image and off-block diagonal values define connections between object detections with compatible classes. The latter are meant to connect detections that could correspond to the same object; since each camera node represents the unique point of view of the corresponding image, no same-class edge is created between camera nodes.

Given the graph $\tilde{\mathcal{G}}$, we encode in each of its nodes $i$ the relevant information from the corresponding image and local map. We define the embedding $\psi(c_{ij}, l_{ij}, b_{ij})=[c_{ij}, \sigma(l_{ij}), \epsilon(b_{ij})]$, where $c_{ij}$ are the coordinates of the object or camera in its local map; $\sigma(l_{ij})$ is the one-hot encoding of the class; and $\epsilon(b_{ij})$ is the bounding box fitted to the detection of the object in the input image. For the camera nodes there is no detection, and we set $\epsilon(b_{ij})=0$.

\subsection{GNN-based Alignment Module}\label{sec:method.alignmentmodule}
We then introduced a three-stage alignment module. \emph{i)} An encoder $\Psi(\cdot)$ composed of a fully connected linear layer with GeLu activation projects the aggregated initial embedding $\psi = [\psi(c_{ij}, l_{ij}, b_{ij})]_{\tilde{V}}$ into a higher-dimensional space, such that $\psi'=\Psi(\psi)\in \mathcal{R}^{F}$, where $F$ is the dimension of the embedding space. \emph{ii)} The projected embedding is fed to a  GNN ($\Xi(\cdot, \cdot)$), producing an updated embedding $\psi''(\cdot)=\Xi(\psi', \tilde{\mathcal{G}})$. The proposed method can be adapted for a wide range of popular GNN,  and a comparative study is available in Section~\ref{sec:experiments}. \emph{iii)} We project into coordinates on a map as $\hat{c}=\Phi(\psi'')\in \mathcal{R}^2$, where $\Phi$ is a decoder based on a fully-connected layer. This outputs the 2D positions of the objects in a common reference frame, \ie the global map. 

The alignment module is trained in supervised fashion, minimizing the linear combination of three different losses: 

\textbf{Euclidean Camera-Object Pose Loss} is the difference between the predicted $\hat{c}_j$, and GT $c^{GT}_j$ pose of object $j$: 
\begin{equation}
    \mathcal{L}_e = \frac{1}{|\tilde{\mathcal{V}}|} \sum_{j=1}^{|\tilde{\mathcal{V}}|} \| \hat{c_l} - c^{GT}_l \|^2_2,
\end{equation}

\textbf{Cross-Map Consistency Loss} is the variance of the $\hat{c_i}$ predicted for detections of object $j$, plus the distance between the mean pose $\mu_j$ and the ground truth $c^{GT}_j$:  
\begin{equation}
    \mathcal{L}_\mu = \frac{1}{T} \sum_{j=1}^T \left( \frac{1}{N_j} \sum_{i=1}^{N_j} \| \hat{c}_i - \hat{\mu}_j \|^2_2 + \|c^{GT}_j - \hat{\mu}_j\|^2_2 \right) 
    \end{equation}
\begin{equation}
    \quad \hat{\mu}_j = \frac{1}{N_j}\sum_{j=1}^{N_j} \hat{c}_j,
\end{equation}
where $T$ is the number of physical objects in the scene and $N_l$ is the number of nodes corresponding to object $j$. This imposes the convergence of pose predicted for the same scene element, by forcing the matched detections to have the same predicted pose, close to the $GT$ location. 

 \textbf{Self-Similarity Loss} measures the consistency between the input local map and the predicted object locations in the global reference system.
For each subgraph $G_i$, the original coordinates in the local map $c^0_{G_i}=\{c_i\}_{i\in G_i}$ and the predicted coordinates in the global reference system $\hat{c}_{G_i}=\{\hat{c}_i\}_{i\in G_i}$ define the self-consistency loss as:
\begin{equation}\label{eq:self-similarity}
    \mathcal{L}_\sigma = \sum_{G_i\in \tilde{G}} \| c^0_{G_i} - (\lambda R \cdot \hat{c}_{G_i} + \tau) \| , \quad 
    \end{equation}
    \begin{equation}
        \quad \lambda, R, \tau = \Theta(c^0_{G_i}, \hat{c}_{G_i})
\end{equation}
where $\Theta$ is a 2D Procrustean alignment algorithm that finds the rigid transformation - defined by a rotation angle $\theta$, a scaling factor $\lambda$ and a translation $\tau$ - that solves the problem:
\begin{align}
    \Theta(c_i,c_j) = & \argmin_{\theta,\tau, \lambda} \| c_i - (\lambda R(\theta) \cdot c_j + \tau) \|.
    \label{eq:align2d}
\end{align}

To solve equation~\eqref{eq:align2d}, we design a differentiable 2D alignment algorithm; implementation and a complete derivation are available in the Supplementary Material. 
This loss reflects the fact that, within each local map, the relative spatial arrangement of the objects should be unchanged, regardless the coordinate system (global or local).

\textbf{Total Loss} uses hyperparameters $\lambda_e, \lambda_\mu$ and $\lambda_\sigma$ to balance the three loss terms:
\begin{equation}\label{eq:losses}
      \mathcal{L} = \lambda_e \mathcal{L}_e + \lambda_\mu \mathcal{L}_\mu +  \lambda_\sigma \mathcal{L}_\sigma,
\end{equation}

Given the intrinsic ambiguity in choosing the reference system of the reconstruction, we expressing the $GT$ always in the reference system of the first subgraph. 
At inference, the 2D coordinates regressed by the Alignment Module $\hat{c}$ are assumed to be the location of the observed objects in a global reference system, as illustrated in Figure~\ref{fig:themethod}.c.

\section{Experiments}\label{sec:experiments}

The MfM problem is new, and there are no readily available datasets and baselines in the literature. Therefore, we construct a dataset from Flatlandia~\cite{toso2023you}, a related dataset for single-view camera localization on object maps in 3DoF. We also generate a synthetic dataset used for ablations on MfM. Both datasets are described in Section~\ref{sec:datasets}.

All models are first tested on the real-world data, to investigate whether the MfM problem is solvable with the proposed approach (Section~\ref{sec:experiments:sfmo}). Then, we use the synthetic dataset to perform ablation tests on the effect of visibility and noise in the local maps on MfM (Section~\ref{sec:experiments:ablations}).

All evaluations report performance in terms of localization accuracy in meters for the predicted camera ($\mu_{c}$) and objects ($\mu_{o}$)  and their respective standard deviations $\sigma_c, \sigma_o$ computed over three runs initialized with different seeds; for each experiment we highlight the \textbf{best} and \underline{second best} result. We also show, where appropriate, the failure rate of the methods.
This is defined as the percentage of scenes where the error is significant \ie greater than $7.5$ meters, or for which it was not possible to generate a reconstruction. The latter case happened, for example, when COLMAP cannot find enough keypoint matches to initialize the SfM reconstruction process. 
As a final remark on evaluation, MfM is also subject to gauge freedom \cite{morris2000uncertainty} as most reconstruction algorithms do, i.e. the global map predicted by MfM is not in the same reference frame of the ground truth data. We, therefore, use the alignment algorithm of Equation~\eqref{eq:align2d} to register the predicted objects and camera locations onto the ground truth global map. This also allows us to report the performance on a metric scale.

\subsection{Datasets}\label{sec:datasets}

\begin{table}[tb]
\footnotesize
\centering
 \begin{tabularx}{\linewidth}{X|c@{\hskip 0.1in}c@{\hskip 0.075in}c@{\hskip 0.075in}c@{\hskip 0.075in}c@{\hskip 0.075in}|c@{\hskip 0.07in}} 
City & Barcelona & Berlin & Lisbon & Vienna & Paris & Avg. \\ [0.5ex]  
\hline & & &&&&\\ [-1.5ex] 
Scenes & 9 & 2 & 22 & 3 & 20 & 11.2\\
Avg. Objects & 18.2 & 70.5 & 27.1 & 65.7 & 17.4 & 39.8\\
Avg. Images & 16.2 & 86 & 31 & 127 & 21.9 & 56.4 \\
Track Length & 5.8 & 4.3 & 4.7 & 5.3 & 4.9 & 5 \\
\end{tabularx}
 \caption{\label{tab:mapo} Statistics on the scenes from extracted from Flatlandia. We report for each city the number of distinct sub-scenes and their average number of objects, images and average track length.
 }
\end{table}

\textbf{\textit{MfM Dataset - based on}~\cite{toso2023you}.}
The Flatlandia~\cite{toso2023you} dataset addresses the problem of 3DoF visual localization, and provides 2D local and reference maps obtained in $20$ locations over $5$ European cities. The dataset provides $20$ reference maps annotated with a total of $2967$ static objects, 
$6.3k$ reference images, and $2k$ query images. For the latter, the dataset provides local maps and their correspondences to the reference maps' objects. 
The local maps are provided in two forms \textit{i)} ground truth: where the position of objects and the camera is based on the global map (i.e. perfect maps); and \textit{ii)} noisy ones obtained using MiDaS~\cite{ranftl2020towards}, a generic monocular depth estimation model, to estimate the layout of detected objects (Depth). These are meant as a noisy limit case. The local maps are obtained using the nominal intrinsic parameters provided by the image metadata, without requiring a calibration step. This is a reasonable approximation, with values within $5\%$ from the intrinsics calibrated using COLMAP.
The dataset's ground truth is obtained by reconstructing the scenes using all data and the COLMAP~\cite{schoenberger2016sfm} SfM approach, the gold standard for reconstruction from multi-view images and a common baseline for scene reconstructions.

We construct a graph where each image from~\cite{toso2023you} is a node. Image pairs are connected if they have detections of at least three objects in common. Then, we extract all connected subgraphs from each graph—sets of images with a sufficient number of matched detections, \ie three, for our specific task. This pre-processing results in two dataset: 

\noindent\textbf{- MfM Large.} We use the entire dataset and produce $56$ sub-graps, with  $32$  images on average per subgraph and $7$ detections per image; the sequences are then split between training ($51\%$), validation ($24\%$) and testing ($25\%$), ensuring no image and no object appears in more than one split.

\noindent\textbf{- MfM Small.} We extract from MfM Large subgraphs of five images with matched detections. This results in $70k$ sequences, evenly split into training, validation and testing. 

Our dataset split addresses the uneven geographical distribution of data in~\cite{toso2023you}, which results in a wide range of number of scenes per city, each with a different amount of images and objects (Table~\ref{tab:mapo}). To provide a more balanced dataset, we distribute the sequences to approach a $50\%$ training, $25\%$ testing and validation split, with a similar variety in the number of images and of objects. This does not prevent generalization, as shown in Section~\ref{sec:experiments:ablations}.

\noindent\textbf{\textit{MfM Synthetic Dataset.}}
For the Synthetic dataset, we generate maps by creating semantic maps with $N_O$ objects uniformly distributed in a box defined within the range $[-1,1]$. Each object is randomly assigned a label from among $N_C$ possible classes. Subsequently, we generate $N_M$ local semantic maps by randomly choosing a point of origin, selecting a subset of objects with each object observed with a probability $\phi$, and choosing a random, compatible viewing direction. This setup offers an ideal testing scenario for proving the concept of the proposed method. In this evaluation, we set $N_C=5$, $N_M=8$ and $N_O=7$, and we provide an ablation of these values in the Supplementary Material.

\subsection{Graph Neural Network Architectures}\label{sec:gnn_models}
We consider different architectural solutions for our \textbf{Alignment module}, considering two categories of GNN models: \textit{i)} Attention-based and \textit{ii)} Non-Attention-based. The attention-based models are designed to assign varying weights to the edges between nodes in the graph. This is achieved through a trainable attention mechanism that assigns larger weights to certain edges while diminishing the significance of others. Attention allows the model to focus on specific relationships within the graph. 
In contrast, the non-attention-based approach uses a more straightforward method for information aggregation. In these models, every edge in the graph is treated based on a pre-defined adjacency matrix, with no dynamic weight training capability. For these experiments, we adopt four different architectures, applying four consecutive layers of the following: \\
\noindent{\textbf{- GCN}~\cite{kipf2016semi}} uses a graph convolutional layer to aggregate information from neighboring nodes of ${G}$, as 
$\psi''=\Xi(\psi', \tilde{\mathcal{G}}) = {\Theta} \sum_{v \in
\mathcal{N}(u) \cup \{ u \}} ({d}_u{d}_v)^{-\frac{1}{2}}{A_{uv}} \mathbf{\psi'}_v
$. Here $A_{uv} \in A$, $\mathcal{N}(u)$ is the neighborhood of node $u$, $d_u$ is its vertex degree, and $\theta$ are the GNN weights.

\noindent\textbf{- GAT}~\cite{brody2021attentive} incorporates an attention mechanism between the node embedding, as 
$
\psi''=\Xi(\psi', \tilde{\mathcal{G}}) = \alpha_{uu}{\Theta}\psi'_u +
\sum_{v \in \mathcal{N}(u)} \alpha_{uv}{\Theta}\psi'_{v}
$, where $\alpha_{uv}$ is the attention mechanism. 

\noindent\textbf{- SuperGAT}~\cite{kim2020find} builds upon the previous method, GAT. The main difference is that SuperGAT employs two types of attention mechanisms, enabling self-supervised learning 
of which edges carry 
the most information between nodes.

\noindent{\textbf{- TransformerGCN}}~\cite{shi-graphtransformer} combines the multi-head attention mechanism of the Transformer\cite{NIPS2017_3f5ee243} and a fusion mechanism,  applied to a standard GCN.
$
\psi''=\Xi(\psi', \tilde{\mathcal{G}}) = \Theta_1 \psi'_u +
\sum_{v \in \mathcal{N}(u)} \alpha_{uv} \Theta_2 \psi'_{v}
$, with $\alpha_{uv}$ computed via multi-head dot product attention.

\subsection{MfM Evaluation}~\label{sec:experiments:sfmo}
The first set of experiments evaluates the MfM solution accuracy on the real-world scenes from \textbf{MfM Dataset}. We consider four different scenarios of decreasing difficulty: \emph{a)} the standard graph formulation as proposed in (Section~\ref{sec:method.graph_representation}) with noisy local maps as input, generated using monocular depth estimation; \emph{b)} the standard graph formulation  with no known matches between the detections in different views but with the ground truth position of each node in its local maps;  \emph{c)} the graph is modified to include the ground truth detection matches, connecting nodes from different subgraphs only if they represent the same object, with noisy local maps as input; \emph{d)} the graph with ground-truth detection matches with the the ground truth local maps.
\begin{table}[h!]
\scriptsize
\begin{tabularx}{\columnwidth}{X|c|ccc}
\hline & \\[-1.5ex]
 & Method & Fail ($\%$) & $\mu_c \pm\sigma_c\;(m)$ & $\mu_o \pm\sigma_o\;(m)$ \\ \hline  & \\[-1.5ex]
\multirow{5}{*}{\rotatebox{90} {Small Scenes}} & COLMAP (Baseline) & 80 & \textbf{2.50} $\pm$ 2.64 & \textbf{1.16} $\pm$ 2.09 \\
 & MfM + GCN & \underline{34} & 3.70$\pm$ 2.12 & 3.82 $\pm$ 1.47 \\
 & MfM + GAT & \textbf{30} & 3.75 $\pm$ 2.17 & 3.58 $\pm$ 1.66 \\
 & MfM + SuperGAT & 36 & \underline{3.58} $\pm$ 2.11 & 3.65 $\pm$ 1.57 \\
 & MfM + Transf.GCN & 37 & 3.73 $\pm$ 2.24 & \underline{3.48} $\pm$ 1.59 \\ \hline & \\[-1.5ex]
\multirow{5}{*}{\rotatebox{90}{Large Scenes}} & COLMAP (Baseline) & 45 & \textbf{1.12} $\pm$ 1.52 & \textbf{1.20} $\pm$ 0.79 \\
 & MfM + GCN & 15 & 4.01 $\pm$ 1.74 & \underline{2.56} $\pm$ 1.15 \\
 & MfM + GAT & \textbf{5} & \underline{3.67} $\pm$ 1.76 & 2.66 $\pm$ 1.28 \\
 & MfM + SuperGAT & \underline{10} & 3.82 $\pm$ 1.68 & 2.74 $\pm$ 1.23 \\
 & MfM + Transf.GCN & \underline{10} & 3.77 $\pm$ 1.78 & 2.72 $\pm$ 1.62 \\ \hline 
\end{tabularx}%
\centering
\caption{\label{tab:baseline_depth} \emph{MfM Dataset Evaluation} (Section~\ref{sec:experiments:sfmo}): Average camera ($\mu_c$) and object ($\mu_o$) error, their standard deviation ($\sigma_c$ and $\sigma_o$), and the failure percentage (Fail) of MfM using 
noisy inputs (Depth Local Maps), compared against standard COLMAP.
}

\end{table}

\begin{table}[]
\scriptsize
\begin{tabularx}{\columnwidth}{X|c|ccc}
\hline & \\[-1.5ex]
 & Method & Fail ($\%$) & $\mu_c \pm\sigma_c\;(m)$ & $\mu_o \pm\sigma_o\;(m)$ \\ \hline  & \\[-1.5ex]
\multirow{5}{*}{\rotatebox{90} {Small Scenes}} & COLMAP (Baseline) & 80 & \textbf{2.50} $\pm$ 2.64 & \textbf{1.16} $\pm$ 2.09 \\
 & MfM + GCN & \textbf{32} & 3.83$\pm$ 2.15 & 3.76 $\pm$ 1.61 \\
 & MfM + GAT & \underline{34} & 3.71 $\pm$ 2.13 & 3.58 $\pm$ 1.46 \\
 & MfM + SuperGAT & 35 & \underline{3.62} $\pm$ 2.21 & 3.67 $\pm$ 1.66 \\
 & MfM + Transf.GCN & 37 & 3.63 $\pm$ 2.15 & \underline{3.51} $\pm$ 1.53 \\ \hline & \\[-1.5ex]
\multirow{5}{*}{\rotatebox{90}{Large Scenes}} & COLMAP (Baseline) & 45 & \textbf{1.12} $\pm$ 1.52 & \textbf{1.20} $\pm$ 0.79 \\
 & MfM + GCN & \underline{10} & 3.98 $\pm$ 1.73 & 2.84 $\pm$ 1.59 \\
 & MfM + GAT & 15 & 3.62 $\pm$ 1.64 & 2.81 $\pm$ 1.87 \\
 & MfM + SuperGAT & \textbf{5} & 4.14 $\pm$ 1.83 & \underline{2.70} $\pm$ 1.49 \\
 & MfM + Transf.GCN & \underline{10} & \underline{3.61} $\pm$ 1.50 & 2.83 $\pm$ 1.39 \\ \hline 
\end{tabularx}%
\centering
\caption{\label{tab:baseline_gt} \emph{MfM Dataset Evaluation} (Section~\ref{sec:experiments:sfmo}): Average camera ($\mu_c$) and object ($\mu_o$) error, their standard deviation ($\sigma_c$ and $\sigma_o$), and the failure percentage (Fail) of MfM using 
perfect inputs (GT Local Maps), compared against standard COLMAP. 
}

\end{table}

\begin{table}[]
\scriptsize
\begin{tabularx}{\columnwidth}{X|c|ccc}
\hline & \\[-1.5ex]
 & Method & Fail ($\%$) & $\mu_c\pm\sigma_c\;(m)$ & $\mu_o\pm\sigma_o\;(m)$ \\ \hline & \\[-1.5ex]
\multirow{5}{*}{\rotatebox{90} {Small Scenes}}& COLMAP (Baseline) & 80 & \textbf{2.50} $\pm$ 2.64 & \textbf{1.16} $\pm$ 2.09 \\
 & MfM + GCN & 35 & 3.72 $\pm$ 2.26 & 4.01 $\pm$1.42 \\
 & MfM + GAT & \underline{30} & 3.60 $\pm$ 2.13 & 3.49 $\pm$1.53 \\
 & MfM + SuperGAT & \textbf{26} & \underline{3.59} $\pm$ 2.07 & \underline{3.44} $\pm$1.63 \\
 & MfM + Transf.GCN & 31 & 3.65 $\pm$ 2.10 & 3.87 $\pm$ 1.39 \\ \hline  & \\[-1.5ex]
\multirow{5}{*}{\rotatebox{90} {Large Scenes}} & COLMAP (Baseline) & 45 & \textbf{1.12} $\pm$ 1.52 & \textbf{1.20} $\pm$ 0.79 \\
 & MfM + GCN & \underline{10} & 4.48 $\pm$ 1.43 & 2.91 $\pm$ 1.63 \\
 & MfM + GAT & \underline{10} & 3.94 $\pm$ 1.74 & \underline{2.58} $\pm$ 1.20 \\
 & MfM + SuperGAT & \textbf{5} & \underline{3.70} $\pm$ 1.98 & 2.67 $\pm$ 1.42 \\
 & MfM + Transf.GCN & \textbf{5} & 3.97 $\pm$ 1.92 & 2.73 $\pm$ 1.39 \\ \hline
\end{tabularx}%

\caption{\label{tab:sfm_known_depth} \emph{MfM Dataset Evaluation with Known Detection Matches} (Section~\ref{sec:experiments:sfmo}): 
Average camera ($\mu_c$) and object ($\mu_o$) error, their standard deviation ($\sigma_c$ and $\sigma_o$), and the failure percentage (Fail) of MfM using 
noisy inputs (Depth Local Maps).
}
\end{table}

\begin{table}[]
\scriptsize
\begin{tabularx}{\columnwidth}{X|c|ccc}
\hline & \\[-1.5ex]
 & Method & Fail ($\%$) & $\mu_c\pm\sigma_c\;(m)$ & $\mu_o\pm\sigma_o\;(m)$ \\ \hline & \\[-1.5ex]
\multirow{5}{*}{\rotatebox{90} {Small Scenes}} & COLMAP (Baseline) & 80 & \textbf{2.50} $\pm$ 2.64 & \textbf{1.16} $\pm$ 2.09 \\
& MfM + GCN & 30 & 3.86 $\pm$ 2.27 & 3.77 $\pm$ 1.56 \\
 & MfM + GAT & \underline{27} & \underline{3.57} $\pm$ 2.05 & 3.53 $\pm$1.53 \\
 & MfM + SuperGAT & \textbf{25} & 3.64 $\pm$ 2.10 & \underline{3.49} $\pm$1.54 \\
 & MfM + Transf.GCN & 31 & 3.86 $\pm$ 2.16 & 3.59$\pm$1.42 \\ \hline & \\[-1.5ex]
\multirow{5}{*}{\rotatebox{90} {Large Scenes}} & COLMAP (Baseline) & 45 & \textbf{1.12} $\pm$ 1.52 & \textbf{1.20} $\pm$ 0.79 \\
 & MfM + GCN & \underline{10} & 3.95 $\pm$ 1.78 & 2.67 $\pm$ 1.42 \\
 & MfM + GAT & \textbf{5} & 3.91 $\pm$ 1.81 & 2.77 $\pm$ 1.33 \\
 & MfM + SuperGAT & \textbf{5} & 3.84 $\pm$ 1.73 & 2.73 $\pm$ 1.45 \\
 & MfM + Transf.GCN & \textbf{5} & \underline{3.66} $\pm$ 1.68 & \underline{2.64} $\pm$ 1.15 \\ \hline
\end{tabularx}%

\caption{\label{tab:sfm_known_gt} \emph{MfM Dataset Evaluation with Known Detection Matches} (Section~\ref{sec:experiments:sfmo}): 
Average camera ($\mu_c$) and object ($\mu_o$) error, their standard deviation ($\sigma_c$ and $\sigma_o$), and the failure percentage (Fail) of MfM using 
perfect inputs (GT Local Maps).
}
\end{table}

\noindent\textbf{COLMAP Baseline.} We compare the performance of MfM against those of COLMAP. Since COLMAP estimates for each scene a set of 3D camera poses and a 3D point cloud, its output cannot be directly compared with the 2D GT data of the MfM dataset, and has to be first projected into 2D points on a horizontal map. After identifying the ground plane of the reconstruction, the 2D camera locations are obtained by directly projecting the camera centers on it; for the object locations, instead, we identify and cluster the 3D points corresponding to all detections of each object, and project the center of the cluster on the ground plane.

\noindent\textbf{MfM Dataset Evaluation.} Table~\ref{tab:baseline_depth} and Table~\ref{tab:baseline_gt} report the results of scenarios \emph{a} and \emph{b} respectively. The COLMAP baseline achieves more accurate localization, as expected. SfM pipelines uses more precise feature point matches together with outlier-free correspondences. This information is not available in the MfM pipeline, which localizes objects on a map rather than computing an accurate 3D point cloud. However, when compared to MfM, COLMAP exhibits a significantly higher failure rate, reaching $45\%$ on MfM Large and a notable $80\%$ on MfM Small. In contrast, the average failure rates for MfM in both  scenarios are around $34\%$ and $10\%$, respectively. This disparity arises due to the sparse nature of input images, leading to insufficient feature matches for initializing the Bundle Adjustment (BA) process that often fails. It is noteworthy that among the MfM models, there is a minimal disparity between the two configurations—when provided with the correct position (scenario \emph{b}) v/s a noisy position (scenario \emph{a}) in the local maps The results highlight the models' resilience to noise in the initial embedding, showcasing comparable performance across scenarios. As a final remark, the proposed approach is significantly more efficient than COLMAP. On both large and small scenes, MfM on average needs $2.8$ ms to estimate the global map, with SAGE being the fastest at $1.8$ ms and SuperGAT being the slowest at $3.5$ ms. In contrast, COLMAP's bundle adjustment requires on average $89.8$ s for MfM large and $0.81$ s for MfM short.

\noindent\textbf{MfM Dataset Evaluation with Known Detection Matches.} Table~\ref{tab:sfm_known_depth} and Table~\ref{tab:sfm_known_gt} report the results of scenarios \emph{c} and \emph{d} respectively, where the graph is modified to include the ground truth detection matches with the correct position and with a noisy position. Looking at the performance, we can see that while the average results are comparable to  Table~\ref{tab:baseline_depth} and Table~\ref{tab:baseline_gt}, the additional information has a positive impact on  the failure rate ($-5\%$), with an average failure rate of $28\%$ for MfM small  and $6.5\%$  for MfM large.
Comparing the two scenarios \emph{c} and \emph{d}, MfM models demonstrate resilience to the noise introduced in the initial embedding. The largest performance gap is observed for the GCN architecture, and is attributed to the model's challenges in mitigating the propagation of incorrect information within the graph. 

\begin{table}[htb!]
\scriptsize
\begin{tabularx}{\columnwidth}{X|cc|cc|cc|cc}

$\Delta_{xy}$ & \multicolumn{2}{c|}{$0.0$m} & \multicolumn{2}{c|}{$2.5$m} & \multicolumn{2}{c|}{$5$m} & \multicolumn{2}{c}{$7.5$m} \\ 
\hline  &&&&&& \\[-1.5ex]
Method & F & $\mu $ & F & $\mu$ & F & $\mu$ & F & $\mu$ \\ \hline   &&&&&& \\[-1.5ex]
MfM + GCN            & 0 & 1.0 & 0 & 0.7 & 1 & 1.0 & 1 & 1.1 \\
MfM + GAT            & 0 & 0.3 & 0 & 0.3 & 0 & 0.6 & 0 & 0.9 \\
MfM + SuperGAT       & 0 & 0.3 & 0 & 0.3 & 0 & 0.7 & 0 & 0.9 \\
MfM + Transf.GCN     & 0 & 0.2 & 0 & 0.2 & 0 & 0.3 & 0 & 0.4 \\ 

\end{tabularx}%
\centering
\caption{\label{tab:mapnoise_ablation_xy} \emph{Effect of Local Map Noise on MfM} (Section~\ref{sec:experiments:ablations}): Failure percentage $F$ and per-object error $\mu\;(m)$ as a function of noise in the local maps' accuracy ($\Delta_{xy}$).}

\end{table}

\begin{table}[htb!]
\scriptsize
\begin{tabularx}{\columnwidth}{X|cc|cc|cc|cc}

$\phi$ & \multicolumn{2}{c|}{$1.0$} & \multicolumn{2}{c|}{$0.75$} & \multicolumn{2}{c|}{$0.5$} & \multicolumn{2}{c}{$0.25$} \\ 
\hline  &&&&&& \\[-1.5ex]
Method & F & $\mu $ & F & $\mu$ & F & $\mu$ & F & $\mu$ \\ \hline   &&&&&& \\[-1.5ex]
MfM + GCN            & 0 & 1.0 & 75 & 5.8 & 97 & 6.4 & 99 & 6.7 \\
MfM + GAT            & 0 & 0.3 & 75 & 5.7 & 97 & 6.4 & 99 & 6.2 \\
MfM + SuperGAT       & 0 & 0.3 & 76 & 5.8 & 97 & 6.0 & 99 & 6.3 \\
MfM + Transf.GCN     & 0 & 0.2 & 75 & 5.7 & 97 & 6.2 & 99 & 4.5 \\  

\end{tabularx}%
\centering
\caption{\label{tab:mapnoise_ablation_vis} \emph{Effect of Visibility on MfM} (Section~\ref{sec:experiments:ablations}): Failure percentage $F$ and per-object error $\mu\;(m)$ as a function of visibility $\phi$.}

\end{table}

\subsection{Ablation}~\label{sec:experiments:ablations}

To investigate the dependence of MfM on the accuracy of the local maps and the density of the graph's edge list, we conduct two experiments on the Synthetic MfM dataset: \textit{i)} varying noise level on the objects' locations in the local maps, and \textit{ii)} changing the visibility, \ie the fraction of the scene's objects observed by each local map. 

We apply displacements in a random direction and an amplitude in the range $[0, \Delta_{xy}]$, with $\Delta_{xy}\in\{0, 2.5, 5, 7.5\}$ meters. The results, reported in Table~\ref{tab:mapnoise_ablation_xy}, show the robust performance of all methods, featuring low failure rates and precise localization despite the injection of noise in the initial map coordinates. Nevertheless, the discernible impact of noise becomes evident as mean errors increase at higher noise levels. The GCN model exhibits a constant slight rise in failure rates, possibly attributable to its relatively lower complexity compared to alternative solutions.

We then test the robustness of MfM to occlusions and other visibility-reducing effects. Table~\ref{tab:mapnoise_ablation_vis} reports performance for different levels of visibility $\phi\in\{1.0, 0.75, 0.50, 0.25\}$, where $\phi$ is the probability of each scene's object being observed by a given camera. With perfect detections ($\phi=1.0$), the mean object localization achieves sub-meter accuracy (maximum error is 0.96 for GCN), with zero failures. However, as visibility decreases, we observe a substantial reduction in localization accuracy. This makes good visibility fundamental to solve MfM, but as long as enough objects are partially observed we can achieve accurate localization, irregardless of camera view. In contrast, SfM requires matching visual features, \ie overlapping fields of view, which is a stricter requirement.

\begin{table}[tb!]
\scriptsize
\centering
\addtolength{\tabcolsep}{-0.1em}
 \begin{tabularx}{\columnwidth}{X| c c c | c c c} 
 &  \multicolumn{3}{c}{MfM+GCN} & \multicolumn{3}{c}{MfM+TransformerGCN}\\
Test Set & Fail & $\mu_c\pm\sigma_c$ & $\mu_o\pm\sigma_o$ & Fail & $\mu_c\pm\sigma_c$ & $\mu_o\pm\sigma_o$ \\ [0.5ex]  
\hline &&&&&&\\[-1.5ex]
 Barcelona & \textbf{0} & 3.9 $\pm$ 1.1 & \textbf{1.7} $\pm$ 0.5 & \textbf{0} & 4.2 $\pm$ 1.4 & \textbf{1.9} $\pm$ 0.6 \\
 Berlin    & \textbf{0} & 5.1 $\pm$ 1.5 & 3.6 $\pm$ 1.4 & \textbf{0} & 5.3 $\pm$ 1.5 & 3.6 $\pm$ 1.2 \\
 Lisbon    & 10 & \textbf{3.5} $\pm$ 2.0 & \underline{2.7} $\pm$ 1.2 & 10 & \textbf{3.3} $\pm$ 1.8 & \underline{2.7} $\pm$ 1.2 \\
 Vienna    & 10 & \textbf{3.5} $\pm$ 1.8 & 3.0 $\pm$ 1.7 & 15 & \underline{3.5} $\pm$ 2.0 & 3.4 $\pm$ 1.7 \\
 Paris     & 15 & \textbf{3.5} $\pm$ 1.8 & 3.0 $\pm$ 1.7 & 10 & \underline{3.5} $\pm$ 2.0 & 3.4 $\pm$ 1.7 \\
Mean       & \underline{9.8} & \underline{3.6} $\pm$ 1.7 & \underline{2.7} $\pm$ 1.4 & \underline{8.3} & 3.6 $\pm$ 1.8 & 2.9 $\pm$ 1.3 \\
\hline &&&&&& \\[-1.5ex]
 Table~\ref{tab:baseline_gt} & 10 & 4.0 $\pm$ 1.7 & 2.8 $\pm$ 1.6 & 10 & 3.6 $\pm$ 1.5 & 2.8 $\pm$ 1.4 \\
\end{tabularx}

 \caption{\label{tab:loo} \emph{Generalization of MfM}. Performance testing on a city and training on the others, 
 reporting the average camera ($\mu_c \pm \sigma_c$) and object ($\mu_o\pm \sigma_o$) error and the fraction of failed scenes (Fail).}
\end{table}

We then test the generalization capabilities of MfM, by performing leave-one-out cross-validation over the imbalanced data of the $5$ cities. Table~\ref{tab:loo} shows the performance of two models - MfM + TransformerGCN and MfM + CGN - tested on sequences from one city and trained on the remaining ones; on average, the weighted mean of these results differs from the values reported in Tab.~\ref{tab:baseline_gt} by only $5.8\%$, confirming that the model can generalize to unseen cities.

\section{Conclusion}

In the paper, we propose a solution to the novel task of MfM, \ie generating 2D object maps by fusing into a global reference frame a set of local 2D semantic maps, representing the partial 2D map as observed from their view point.  This initial method provides a first step towards developing tools to automatically generate annotations on 2D maps from uncalibrated images, without having to generate a 3D reconstruction of the scene or establish matches between the images. 
We shown how the proposed approach, MfM, can provide accurate maps even in the presence of limited information, such as noisy input maps 
and no availability of cross-view match between the detected objects. 
Even in these challenging scenarios, MfM provides an average localization within GPS accuracy. Moreover, when applied to sequences of very sparse images (e.g. five) with large viewpoint changes, MfM achieves approximately three times lower failure rate than COLMAP.

The main limitation of the approach, as demonstrated in the ablation experiments, lies in its dependency on 
objects' covisibility among the multi-view images. Based on empirical estimations, every image must share at least three detections with another view. Images containing only classes not observed in the other images will result in disconnected subgraphs in $\tilde{\mathcal{G}}$, making the alignment impossible.

Future work will expand the approach to predict accurate detection matches across maps, to improve the information aggregation capabilities of graph-based learning models. This will allow extending the approach to tasks like multi-view object localization.
Additionally, depth estimation is not the most accurate way for generating top-view maps, and was used to provide a lower-accuracy limit; future work will employ more sophisticated approaches, like generating segmented BEV maps from each image. 

\subsubsection*{Acknowledgments}
This work is supported by PNRR MUR Project Cod. PE0000013 "Future Artificial Intelligence Research (hereafter FAIR)" - CUP J53C22003010006, and by the European Union's Horizon research and innovation programme "DCitizens" (grant agreement No 101079116) and "Bauhaus of the Seas Sails" (grant agreement No 101079995).

{
    \small
    \bibliographystyle{ieeenat_fullname}
    \bibliography{main}

\begin{thebibliography}{63}
\providecommand{\natexlab}[1]{#1}
\providecommand{\url}[1]{\texttt{#1}}
\expandafter\ifx\csname urlstyle\endcsname\relax
  \providecommand{\doi}[1]{doi: #1}\else
  \providecommand{\doi}{doi: \begingroup \urlstyle{rm}\Url}\fi

\bibitem[Agarwal et~al.(2009)Agarwal, Snavely, Simon, Seitz, and
  Szeliski]{romeinaday}
Sameer Agarwal, Noah Snavely, Ian Simon, Steven~M. Seitz, and Richard Szeliski.
\newblock Building rome in a day.
\newblock In \emph{{ICCV} 2009}, pages 72--79, 2009.

\bibitem[Armeni et~al.(2019)Armeni, He, Gwak, Zamir, Fischer, Malik, and
  Savarese]{armeni20193d}
Iro Armeni, Zhi-Yang He, JunYoung Gwak, Amir~R Zamir, Martin Fischer, Jitendra
  Malik, and Silvio Savarese.
\newblock 3d scene graph: A structure for unified semantics, 3d space, and
  camera.
\newblock In \emph{Proceedings of the IEEE/CVF international conference on
  computer vision}, pages 5664--5673, 2019.

\bibitem[Bandil et~al.(2021)Bandil, Girdhar, Chau, Ali, Hendawi, Govind, Cao,
  and Song]{bandil2021geodart}
Ayush Bandil, Vaishali Girdhar, Hieu Chau, Mohamed Ali, Abdeltawab Hendawi,
  Harsh Govind, Peiwei Cao, and Ashley Song.
\newblock Geodart: A system for discovering maps discrepancies.
\newblock In \emph{2021 IEEE 37th International Conference on Data Engineering
  (ICDE)}, pages 2535--2546. IEEE, 2021.

\bibitem[Benbihi et~al.(2022)Benbihi, Pradalier, and Chum]{benbihi2022}
A. Benbihi, C. Pradalier, and O. Chum.
\newblock Object-guided day-night visual localization in urban scenes.
\newblock In \emph{2022 26th International Conference on Pattern Recognition
  (ICPR)}, pages 3786--3793, Los Alamitos, CA, USA, 2022. IEEE Computer
  Society.

\bibitem[Brody et~al.(2021)Brody, Alon, and Yahav]{brody2021attentive}
Shaked Brody, Uri Alon, and Eran Yahav.
\newblock How attentive are graph attention networks?
\newblock In \emph{International Conference on Learning Representations}, 2021.

\bibitem[Brynte et~al.(2023)Brynte, Iglesias, Olsson, and
  Kahl]{Brynte2023LearningSW}
Lucas Brynte, Jos{\'e}~Pedro Iglesias, Carl Olsson, and Fredrik Kahl.
\newblock Learning structure-from-motion with graph attention networks.
\newblock \emph{ArXiv}, abs/2308.15984, 2023.

\bibitem[Chen et~al.(2024)Chen, Li, Liu, Shen, Li, and
  Tian]{chen2024integrating}
Yanming Chen, Guoli Li, Xiaoqiang Liu, Yueqian Shen, Jia Li, and Qin Tian.
\newblock Integrating openstreetmap tags for efficient lidar point cloud
  classification using graph neural networks.
\newblock \emph{International Journal of Digital Earth}, 17\penalty0
  (1):\penalty0 2297946, 2024.

\bibitem[Contributors(2017)]{osm}
OpenStreetMap Contributors.
\newblock Planet dump retrieved from https://planet.osm.org, 2017.

\bibitem[Crocco et~al.(2016)Crocco, Rubino, and Del~Bue]{crocco2016structure}
Marco Crocco, Cosimo Rubino, and Alessio Del~Bue.
\newblock Structure from motion with objects.
\newblock In \emph{Proceedings of the IEEE Conference on Computer Vision and
  Pattern Recognition}, pages 4141--4149, 2016.

\bibitem[Dhamo et~al.(2021)Dhamo, Manhardt, Navab, and
  Tombari]{graph2scene2021}
Helisa Dhamo, Fabian Manhardt, Nassir Navab, and Federico Tombari.
\newblock Graph-to-3d: End-to-end generation and manipulation of 3d scenes
  using scene graphs.
\newblock In \emph{IEEE International Conference on Computer Vision (ICCV)},
  2021.

\bibitem[Duchi et~al.(2011)Duchi, Hazan, and Singer]{duchi2011adaptive}
John Duchi, Elad Hazan, and Yoram Singer.
\newblock Adaptive subgradient methods for online learning and stochastic
  optimization.
\newblock \emph{Journal of machine learning research}, 12\penalty0 (7), 2011.

\bibitem[Durrant-Whyte and Bailey(2006)]{1638022}
H. Durrant-Whyte and T. Bailey.
\newblock Simultaneous localization and mapping: part i.
\newblock \emph{IEEE Robotics and Automation Magazine}, 13\penalty0
  (2):\penalty0 99--110, 2006.

\bibitem[Elich et~al.(2023)Elich, Armeni, Oswald, Pollefeys, and
  Stueckler]{cathrin}
Cathrin Elich, Iro Armeni, Martin~R. Oswald, Marc Pollefeys, and Joerg
  Stueckler.
\newblock Learning-based relational object matching across views.
\newblock In \emph{2023 IEEE International Conference on Robotics and
  Automation (ICRA)}, pages 1--7, 2023.

\bibitem[Fiorini et~al.(2023)Fiorini, Coniglio, Ciavotta, and
  Messina]{fiorini2023sigmanet}
Stefano Fiorini, Stefano Coniglio, Michele Ciavotta, and Enza Messina.
\newblock Sigmanet: One laplacian to rule them all.
\newblock In \emph{Proceedings of the AAAI Conference on Artificial
  Intelligence}, 2023.

\bibitem[Fiorini et~al.(2024)Fiorini, Coniglio, Ciavotta, and
  Messina]{fiorini2024graph}
Stefano Fiorini, Stefano Coniglio, Michele Ciavotta, and Enza Messina.
\newblock Graph learning in 4d: A quaternion-valued laplacian to enhance
  spectral gcns.
\newblock In \emph{Proceedings of the AAAI Conference on Artificial
  Intelligence}, pages 12006--12015, 2024.

\bibitem[Frost et~al.(2018)Frost, Prisacariu, and Murray]{8353862}
Duncan Frost, Victor Prisacariu, and David Murray.
\newblock Recovering stable scale in monocular slam using object-supplemented
  bundle adjustment.
\newblock \emph{IEEE Transactions on Robotics}, 34\penalty0 (3):\penalty0
  736--747, 2018.

\bibitem[Fuchs et~al.(2020)Fuchs, Worrall, Fischer, and Welling]{fuchs2020se}
Fabian Fuchs, Daniel Worrall, Volker Fischer, and Max Welling.
\newblock Se(3)-transformers: 3d roto-translation equivariant attention
  networks.
\newblock \emph{Advances in neural information processing systems},
  33:\penalty0 1970--1981, 2020.

\bibitem[Garg et~al.(2021)Garg, Dhamo, Farshad, Musatian, Navab, and
  Tombari]{garg2021unconditional}
Sarthak Garg, Helisa Dhamo, Azade Farshad, Sabrina Musatian, Nassir Navab, and
  Federico Tombari.
\newblock Unconditional scene graph generation.
\newblock In \emph{Proceedings of the IEEE/CVF International Conference on
  Computer Vision}, pages 16362--16371, 2021.

\bibitem[Gay et~al.(2017)Gay, Rubino, Bansal, and Del~Bue]{Gay_2017_ICCV}
Paul Gay, Cosimo Rubino, Vaibhav Bansal, and Alessio Del~Bue.
\newblock Probabilistic structure from motion with objects (psfmo).
\newblock In \emph{Proceedings of the IEEE International Conference on Computer
  Vision (ICCV)}, 2017.

\bibitem[Gay et~al.(2019)Gay, Stuart, and Del~Bue]{gay2019visual}
Paul Gay, James Stuart, and Alessio Del~Bue.
\newblock Visual graphs from motion (vgfm): Scene understanding with object
  geometry reasoning.
\newblock In \emph{Computer Vision--ACCV 2018: 14th Asian Conference on
  Computer Vision, Perth, Australia, December 2--6, 2018, Revised Selected
  Papers, Part III 14}, pages 330--346. Springer, 2019.

\bibitem[Giuliari et~al.(2022)Giuliari, Skenderi, Cristani, Wang, and
  Del~Bue]{giuliari2022spatial}
Francesco Giuliari, Geri Skenderi, Marco Cristani, Yiming Wang, and Alessio
  Del~Bue.
\newblock Spatial commonsense graph for object localisation in partial scenes.
\newblock In \emph{Proceedings of the IEEE/CVF Conference on Computer Vision
  and Pattern Recognition}, pages 19518--19527, 2022.

\bibitem[Giuliari et~al.(2024)Giuliari, Scarpellini, Fiorini, James, Morerio,
  Wang, and Del~Bue]{giuliari2023positional}
Francesco Giuliari, Gianluca Scarpellini, Stefano Fiorini, Stuart James, Pietro
  Morerio, Yiming Wang, and Alessio Del~Bue.
\newblock Positional diffusion: Graph-based diffusion models for set ordering.
\newblock \emph{Pattern Recognition Letters}, 2024.

\bibitem[Gu et~al.(2019)Gu, Zhao, Lin, Li, Cai, and Ling]{gu2019scene}
Jiuxiang Gu, Handong Zhao, Zhe Lin, Sheng Li, Jianfei Cai, and Mingyang Ling.
\newblock Scene graph generation with external knowledge and image
  reconstruction.
\newblock In \emph{Proceedings of the IEEE/CVF conference on computer vision
  and pattern recognition}, pages 1969--1978, 2019.

\bibitem[He et~al.(2020)He, Bastani, Jagwani, Park, Abbar, Alizadeh,
  Balakrishnan, Chawla, Madden, and Sadeghi]{he2020roadtagger}
Songtao He, Favyen Bastani, Satvat Jagwani, Edward Park, Sofiane Abbar,
  Mohammad Alizadeh, Hari Balakrishnan, Sanjay Chawla, Samuel Madden, and
  Mohammad~Amin Sadeghi.
\newblock Roadtagger: Robust road attribute inference with graph neural
  networks.
\newblock In \emph{Proceedings of the AAAI Conference on Artificial
  Intelligence}, pages 10965--10972, 2020.

\bibitem[Huang et~al.(2022)Huang, Qiu, Yu, and Lu]{huang2022msen}
Zongcai Huang, Peiyuan Qiu, Li Yu, and Feng Lu.
\newblock Msen-grp: A geographic relations prediction model based on
  multi-layer similarity enhanced networks for geographic relations completion.
\newblock \emph{ISPRS International Journal of Geo-Information}, 11\penalty0
  (9):\penalty0 493, 2022.

\bibitem[Kim and Oh(2020)]{kim2020find}
Dongkwan Kim and Alice Oh.
\newblock How to find your friendly neighborhood: Graph attention design with
  self-supervision.
\newblock In \emph{International Conference on Learning Representations}, 2020.

\bibitem[Kipf and Welling(2017)]{kipf2016semi}
Thomas.~N. Kipf and Max Welling.
\newblock Semi-supervised classification with graph convolutional networks.
\newblock In \emph{5th International Conference on Learning Representations,
  ICLR 2017 - Conference Track Proceedings}, 2017.

\bibitem[Kirillov et~al.(2019)Kirillov, He, Girshick, Rother, and
  Dollar]{Kirillov_2019_CVPR}
Alexander Kirillov, Kaiming He, Ross Girshick, Carsten Rother, and Piotr
  Dollar.
\newblock Panoptic segmentation.
\newblock In \emph{Proceedings of the IEEE/CVF Conference on Computer Vision
  and Pattern Recognition (CVPR)}, 2019.

\bibitem[Li et~al.(2023)Li, Liu, Cai, Chen, and Chen]{li2023unleashing}
Guoli Li, Xiaoqiang Liu, Xinyu Cai, Yao Chen, and Yanming Chen.
\newblock Unleashing the power of openstreetmap tags: a graph neural network
  approach for efficient lidar point cloud classification.
\newblock In \emph{International Conference on Remote Sensing, Mapping, and
  Geographic Systems (RSMG 2023)}, pages 734--739. SPIE, 2023.

\bibitem[Li et~al.(2021)Li, DeTone, Chen, Vo, Reid, Rezatofighi, Sweeney,
  Straub, and Newcombe]{Li_2021_ICCV}
Kejie Li, Daniel DeTone, Yu~Fan~(Steven) Chen, Minh Vo, Ian Reid, Hamid
  Rezatofighi, Chris Sweeney, Julian Straub, and Richard Newcombe.
\newblock Odam: Object detection, association, and mapping using posed rgb
  video.
\newblock In \emph{Proceedings of the IEEE/CVF International Conference on
  Computer Vision (ICCV)}, pages 5998--6008, 2021.

\bibitem[Liu et~al.(2020)Liu, Ong, and Chen]{liu2020graphsage}
Jielun Liu, Ghim~Ping Ong, and Xiqun Chen.
\newblock Graphsage-based traffic speed forecasting for segment network with
  sparse data.
\newblock \emph{IEEE Transactions on Intelligent Transportation Systems},
  23\penalty0 (3):\penalty0 1755--1766, 2020.

\bibitem[Ma et~al.(2023)Ma, Zhang, and Han]{electronics12245017}
Zhiguo Ma, Yutong Zhang, and Meng Han.
\newblock Predicting maps using in-vehicle cameras for data-driven intelligent
  transport.
\newblock \emph{Electronics}, 12\penalty0 (24), 2023.

\bibitem[Morris et~al.(2000)Morris, Kanatani, and
  Kanade]{morris2000uncertainty}
Daniel~D Morris, Kenichi Kanatani, and Takeo Kanade.
\newblock Uncertainty modeling for optimal structure from motion.
\newblock In \emph{Vision Algorithms: Theory and Practice: International
  Workshop on Vision Algorithms Corfu, Greece, September 21--22, 1999
  Proceedings}, pages 200--217. Springer, 2000.

\bibitem[Moulon et~al.(2016)Moulon, Monasse, Perrot, and
  Marlet]{moulon2016openmvg}
Pierre Moulon, Pascal Monasse, Romuald Perrot, and Renaud Marlet.
\newblock Open{MVG}: Open multiple view geometry.
\newblock In \emph{International Workshop on Reproducible Research in Pattern
  Recognition}, pages 60--74. Springer, 2016.

\bibitem[N.~Samano(2020)]{noe2020eccv}
A.~Calway N.~Samano, M~Zhou.
\newblock You are here: Geolocation by embedding maps and images.
\newblock In \emph{In Proc. of the European Conference on Computer Vision
  (ECCV)}, 2020.

\bibitem[Nassar et~al.(2019)Nassar, Lef{\`e}vre, and
  Wegner]{nassar2019simultaneous}
Ahmed~Samy Nassar, S{\'e}bastien Lef{\`e}vre, and Jan~Dirk Wegner.
\newblock Simultaneous multi-view instance detection with learned geometric
  soft-constraints.
\newblock In \emph{Proceedings of the IEEE/CVF international conference on
  computer vision}, pages 6559--6568, 2019.

\bibitem[Nassar et~al.(2020)Nassar, D’aronco, Lef{\`e}vre, and
  Wegner]{nassar2020geograph}
Ahmed~Samy Nassar, Stefano D’aronco, S{\'e}bastien Lef{\`e}vre, and Jan~D
  Wegner.
\newblock Geograph: graph-based multi-view object detection with geometric cues
  end-to-end.
\newblock In \emph{Computer Vision--ECCV 2020: 16th European Conference,
  Glasgow, UK, August 23--28, 2020, Proceedings, Part VII 16}, pages 488--504.
  Springer, 2020.

\bibitem[Nicholson et~al.(2018)Nicholson, Milford, and
  Sunderhauf]{Nicholson_2018_CVPR_Workshops}
Lachlan Nicholson, Michael Milford, and Niko Sunderhauf.
\newblock Quadricslam: Dual quadrics as slam landmarks.
\newblock In \emph{Proceedings of the IEEE Conference on Computer Vision and
  Pattern Recognition (CVPR) Workshops}, 2018.

\bibitem[Nicholson et~al.(2019)Nicholson, Milford, and Sünderhauf]{8440105}
Lachlan Nicholson, Michael Milford, and Niko Sünderhauf.
\newblock Quadricslam: Dual quadrics from object detections as landmarks in
  object-oriented slam.
\newblock \emph{IEEE Robotics and Automation Letters}, 4\penalty0 (1):\penalty0
  1--8, 2019.

\bibitem[Parkhiya et~al.(2018)Parkhiya, Khawad, Murthy, Bhowmick, and
  Krishna]{8460816}
Parv Parkhiya, Rishabh Khawad, J.~Krishna Murthy, Brojeshwar Bhowmick, and
  K.~Madhava Krishna.
\newblock Constructing category-specific models for monocular object-slam.
\newblock In \emph{2018 IEEE International Conference on Robotics and
  Automation (ICRA)}, pages 4517--4524, 2018.

\bibitem[Purkait et~al.(2020)Purkait, Chin, and Reid]{purkait2020neurora}
Pulak Purkait, Tat-Jun Chin, and Ian Reid.
\newblock Neurora: Neural robust rotation averaging.
\newblock In \emph{European Conference on Computer Vision}, pages 137--154.
  Springer, 2020.

\bibitem[Qin et~al.(2020)Qin, Xu, Kang, and Kwan]{qin2020graph}
Kun Qin, Yuanquan Xu, Chaogui Kang, and Mei-Po Kwan.
\newblock A graph convolutional network model for evaluating potential
  congestion spots based on local urban built environments.
\newblock \emph{Transactions in GIS}, 24\penalty0 (5):\penalty0 1382--1401,
  2020.

\bibitem[Ranftl et~al.(2020)Ranftl, Lasinger, Hafner, Schindler, and
  Koltun]{ranftl2020towards}
Ren{\'e} Ranftl, Katrin Lasinger, David Hafner, Konrad Schindler, and Vladlen
  Koltun.
\newblock Towards robust monocular depth estimation: Mixing datasets for
  zero-shot cross-dataset transfer.
\newblock \emph{IEEE transactions on pattern analysis and machine
  intelligence}, 44\penalty0 (3):\penalty0 1623--1637, 2020.

\bibitem[Ravichandran et~al.(2022)Ravichandran, Peng, Hughes, Griffith, and
  Carlone]{Ravichandran2022Scenenavigation}
Zachary Ravichandran, Lisa Peng, Nathan Hughes, J.~Daniel Griffith, and Luca
  Carlone.
\newblock Hierarchical representations and explicit memory: Learning effective
  navigation policies on 3d scene graphs using graph neural networks.
\newblock In \emph{2022 International Conference on Robotics and Automation
  (ICRA)}, pages 9272--9279, 2022.

\bibitem[Rubino et~al.(2018)Rubino, Crocco, and Del~Bue]{7919240}
Cosimo Rubino, Marco Crocco, and Alessio Del~Bue.
\newblock 3d object localisation from multi-view image detections.
\newblock \emph{IEEE Transactions on Pattern Analysis and Machine
  Intelligence}, 40\penalty0 (6):\penalty0 1281--1294, 2018.

\bibitem[Saha et~al.(2022)Saha, Mendez, Russell, and Bowden]{Saha2022CVPR}
Avishkar Saha, Oscar Mendez, Chris Russell, and Richard Bowden.
\newblock ''the pedestrian next to the lamppost'' adaptive object graphs for
  better instantaneous mapping<br/>.
\newblock In \emph{IEEE/CVF Conference on Computer Vision and Pattern
  Recognition (CVPR)}, pages 19528--19537, 2022.

\bibitem[Sarlin et~al.(2023{\natexlab{a}})Sarlin, DeTone, Yang, Avetisyan,
  Straub, Malisiewicz, Bulo, Newcombe, Kontschieder, and
  Balntas]{sarlin2023orienternet}
Paul-Edouard Sarlin, Daniel DeTone, Tsun-Yi Yang, Armen Avetisyan, Julian
  Straub, Tomasz Malisiewicz, Samuel~Rota Bulo, Richard Newcombe, Peter
  Kontschieder, and Vasileios Balntas.
\newblock {OrienterNet: Visual Localization in 2D Public Maps with Neural
  Matching}.
\newblock In \emph{CVPR}, 2023{\natexlab{a}}.

\bibitem[Sarlin et~al.(2023{\natexlab{b}})Sarlin, Trulls, Pollefeys, Hosang,
  and Lynen]{sarlin2023snap}
Paul-Edouard Sarlin, Eduard Trulls, Marc Pollefeys, Jan Hosang, and Simon
  Lynen.
\newblock {SNAP: Self-Supervised Neural Maps for Visual Positioning and
  Semantic Understanding}.
\newblock In \emph{NeurIPS}, 2023{\natexlab{b}}.

\bibitem[Scarpellini et~al.(2024)Scarpellini, Fiorini, Giuliari, Morerio, and
  Bue]{scarpellini2024diffassemble}
Gianluca Scarpellini, Stefano Fiorini, Francesco Giuliari, Pietro Morerio, and
  Alessio~Del Bue.
\newblock Diffassemble: A unified graph-diffusion model for 2d and 3d
  reassembly, 2024.

\bibitem[Schonberger and Frahm(2016)]{Schonberger_2016_CVPR}
Johannes~L. Schonberger and Jan-Michael Frahm.
\newblock Structure-from-motion revisited.
\newblock In \emph{Proceedings of the IEEE Conference on Computer Vision and
  Pattern Recognition (CVPR)}, 2016.

\bibitem[Sch\"{o}nberger and Frahm(2016)]{schoenberger2016sfm}
Johannes~Lutz Sch\"{o}nberger and Jan-Michael Frahm.
\newblock Structure-from-motion revisited.
\newblock In \emph{Conference on Computer Vision and Pattern Recognition
  (CVPR)}, 2016.

\bibitem[Shi et~al.(2021)Shi, Huang, Feng, Zhong, Wang, and
  Sun]{shi-graphtransformer}
Yunsheng Shi, Zhengjie Huang, Shikun Feng, Hui Zhong, Wenjing Wang, and Yu Sun.
\newblock Masked label prediction: Unified message passing model for
  semi-supervised classification.
\newblock In \emph{IJCAI}, 2021.

\bibitem[Shi et~al.(2023)Shi, Wu, Perincherry, Vora, and Li]{Shi_2023_ICCV}
Yujiao Shi, Fei Wu, Akhil Perincherry, Ankit Vora, and Hongdong Li.
\newblock Boosting 3-dof ground-to-satellite camera localization accuracy via
  geometry-guided cross-view transformer.
\newblock In \emph{Proceedings of the IEEE/CVF International Conference on
  Computer Vision (ICCV)}, pages 21516--21526, 2023.

\bibitem[Taiana et~al.(2022)Taiana, Toso, James, and Bue]{posernet_eccv2022}
Matteo Taiana, Matteo Toso, Stuart James, and Alessio~Del Bue.
\newblock Posernet: Refining relative camera poses exploiting object
  detections.
\newblock In \emph{Proceedings of the European Conference on Computer Vision
  ({ECCV})}, 2022.

\bibitem[Toft et~al.(2018)Toft, Stenborg, Hammarstrand, Brynte, Pollefeys,
  Sattler, and Kahl]{Toft_2018_ECCV}
Carl Toft, Erik Stenborg, Lars Hammarstrand, Lucas Brynte, Marc Pollefeys,
  Torsten Sattler, and Fredrik Kahl.
\newblock Semantic match consistency for long-term visual localization.
\newblock In \emph{Proceedings of the European Conference on Computer Vision
  (ECCV)}, 2018.

\bibitem[Toso et~al.(2023)Toso, Taiana, James, and Del~Bue]{toso2023you}
Matteo Toso, Matteo Taiana, Stuart James, and Alessio Del~Bue.
\newblock You are here! finding position and orientation on a 2d map from a
  single image: The flatlandia localization problem and dataset.
\newblock In \emph{arXiv preprint arXiv:2304.06373}, 2023.

\bibitem[Vaswani et~al.(2017)Vaswani, Shazeer, Parmar, Uszkoreit, Jones, Gomez,
  Kaiser, and Polosukhin]{NIPS2017_3f5ee243}
Ashish Vaswani, Noam Shazeer, Niki Parmar, Jakob Uszkoreit, Llion Jones,
  Aidan~N Gomez, \L~ukasz Kaiser, and Illia Polosukhin.
\newblock Attention is all you need.
\newblock In \emph{Advances in Neural Information Processing Systems}. Curran
  Associates, Inc., 2017.

\bibitem[Vojir et~al.(2020)Vojir, Budvytis, and Cipolla]{Vojir_2020_ACCV}
Tomas Vojir, Ignas Budvytis, and Roberto Cipolla.
\newblock Efficient large-scale semantic visual localization in 2d maps.
\newblock In \emph{Proceedings of the Asian Conference on Computer Vision
  (ACCV)}, 2020.

\bibitem[Xue et~al.(2022)Xue, Budvytis, Reino, and Cipolla]{Xue_2022_CVPR}
Fei Xue, Ignas Budvytis, Daniel~Olmeda Reino, and Roberto Cipolla.
\newblock Efficient large-scale localization by global instance recognition.
\newblock In \emph{Proceedings of the IEEE/CVF Conference on Computer Vision
  and Pattern Recognition (CVPR)}, pages 17348--17357, 2022.

\bibitem[Yew and Lee(2021)]{yew2020-RobustSync}
Zi~Jian Yew and Gim~Hee Lee.
\newblock Learning iterative robust transformation synchronization.
\newblock In \emph{International Conference on 3D Vision (3DV)}, 2021.

\bibitem[Yin et~al.(2020)Yin, Varadarajan, Wang, Wang, Sahrawat, Zimmermann,
  and Ng]{yin2020multi}
Yifang Yin, Jagannadan Varadarajan, Guanfeng Wang, Xueou Wang, Dhruva Sahrawat,
  Roger Zimmermann, and See-Kiong Ng.
\newblock A multi-task learning framework for road attribute updating via joint
  analysis of map data and gps traces.
\newblock In \emph{Proceedings of The Web Conference 2020}, pages 2662--2668,
  2020.

\bibitem[Zhou et~al.(2021)Zhou, Chen, Samano, Stachniss, and Calway]{9635972}
Mengjie Zhou, Xieyuanli Chen, Noe Samano, Cyrill Stachniss, and Andrew Calway.
\newblock Efficient localisation using images and {OpenStreetMaps}.
\newblock In \emph{2021 IEEE/RSJ International Conference on Intelligent Robots
  and Systems (IROS)}, pages 5507--5513, 2021.

\bibitem[Zins et~al.(2020)Zins, Simon, and Berger]{zinshal02975379}
Matthieu Zins, Gilles Simon, and Marie-Odile Berger.
\newblock {3D-Aware Ellipse Prediction for Object-Based Camera Pose
  Estimation}.
\newblock In \emph{{3DV 2020 - International Virtual Conference on 3D Vision}},
  Fukuoka, Japan, 2020.

\end{thebibliography}
}
\newpage
\appendix

\begin{figure*}[t!]
        \centering
        \includegraphics[width=\linewidth, trim={0 1cm 2.5cm 3cm}, clip] {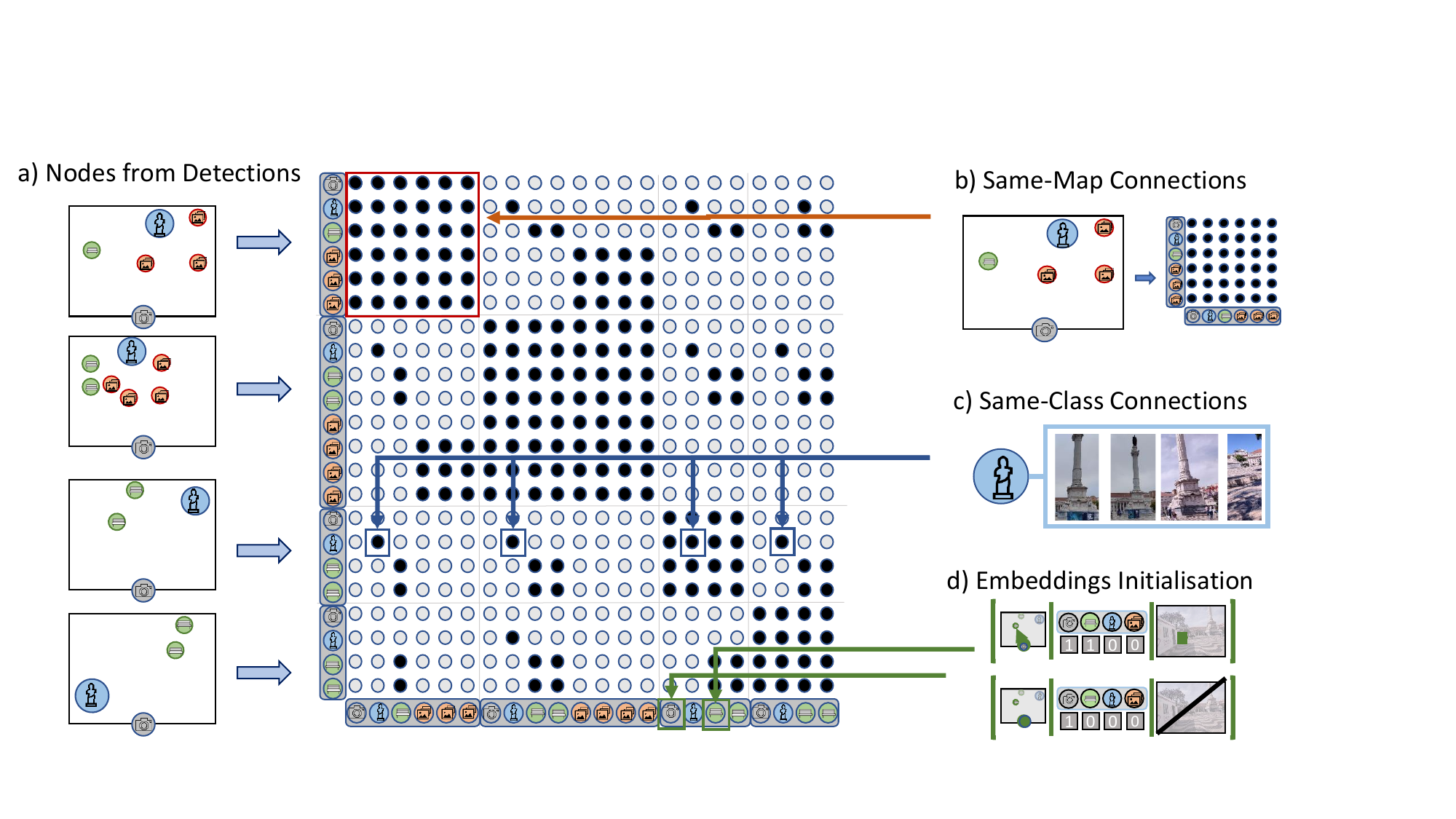}
    \caption{\emph{Graph Formulation for the MfM problem}. Given a set of local maps, we {\em a)} assign to each map annotation a node in the graph and {\em b)} draw intra-map edges to generate a complete subgraph for each map. We then {\em c)} draw inter-map edges connecting detections matched to the same object. Finally, we {\em d)} assign to each node of the graph an embedding defined by concatenating the detection's coordinates in the corresponding local map, the one-hot encoding of its semantic class, and the location of a bounding box fitted to the segmentation mask.}
    \label{fig:thegraph}
\end{figure*}

\section{Derivation of the 2D Alignment Algorithm}\label{sec:procrustes2D}

In Section 3.3 of the main paper, we introduce a \textit{Self-Consistency Loss} that requires finding the optimal transformation aligning two maps, \ie two sets of 2D points, in the same reference frame. To train the model, we need this loss to be differentiable, with stable gradients and efficient to compute. 

To satisfy these requirements, we introduce a novel alignment algorithm constrained to a 2D surface and inspired by the Procrustes algorithm. Consider two 2D points $c_i$ and $c_j$ where $c_i, c_j\in \mathcal{R}^{N\times 2}$, with the same number of elements and sorted such that the $n$-th element of $c_i$ and $c_j$ represent the same object but in different reference frames. These 2D point arrangements can be aligned by finding the rigid transformation defined by a rotation angle $\theta$, a scaling factor $\lambda$, and a translation $\tau$ that solves the problem:
\begin{align}
    \Theta(c_i,c_j) = & \argmin_{\theta,\tau, \lambda} \| c_i - (\lambda R(\theta) \cdot c_j + \tau) \|.
    \label{eq:align2d}
\end{align}

To find this transformation, we first center and 
normalize the two point clouds, to reduce the alignment problem to estimating a rotation around the cloud's centers:
\begin{align}
    \tau_i = & \frac{1}{N}\sum c_i && \tau_j = \frac{1}{N}\sum c_j \\ 
    \lambda_i = & \|c_i - \mu_i\| && \lambda_j = \|c_j - \mu_j\| \\
    c'_i = & \frac{c_i - \tau_i}{\lambda_i} && c'_j = \frac{c_j - \tau_j}{\lambda_j}. 
\end{align}

We can then compare the angular distribution of points in $c'_i$ and $c'_j$, to estimate the rotation angle $\theta_{ji}$ that best aligns them. The orientation of point $l$ in the point cloud $c'_i$, $\theta_l^i$, can be obtained as $\gamma_l^i=\|c^l_i\|^{-1}c^l_i=[\cos(\theta_l^i), \sin(\theta_l^i)]$, \ie by normalizing the vector connecting the point to the center of the cloud $\mu_k$. The two angular distributions are found as $\gamma_i = \left\{ [\cos(\theta_l^i), \sin(\theta_l^i)] \right\}_{l \in c'_i}$ and $\gamma_j = \left\{ [\cos(\theta_l^j), \sin(\theta_l^j)] \right\}_{l \in c'_j}$.

Due to the possible presence of noise in the input 2D point clouds, we compute the average angular distance between pairs of matched points to find the rotation angle $\theta_{ji}$ mapping the two distributions. This is performed using the Prosthaphaeresis formulas, which can be computed as follows:

\begin{align}
    \sin{\theta_{ji}} = & \frac{1}{N}\sum_{l} (\sin{\theta_j^l}\cos{\theta_i^l} - \cos{\theta_j^l}\sin{\theta_i^l}) \\
    \cos{\theta_{ji}} = & \frac{1}{N}\sum_{l} (\cos{\theta_j^l}\cos{\theta_i^l} - \sin{\theta_j^l}\sin{\theta_i^l}) \\
    R(\theta_{ji}) = & \begin{bmatrix} \cos{\theta_{ji}} & \sin{\theta_{ji}} \\ -\sin{\theta_{ji}} & \cos{\theta_{ji}} \end{bmatrix}.
\end{align}

The rotation $R(\theta_{ji})$, the scaling factor $\lambda_i$ and the translation $\tau_i$ then define the optimal transformation mapping $c_j$ onto $c_i$, \ie $c_i \approx \lambda_i R(\theta_{ji}) \cdot c_j + \tau_i$.

\section{Graph Formulation for the MfM Problem}
\begin{algorithm}[htb]
  \KwData{Images $I_i$, Objects $O_i$, Cameras $C_i$, Objects' Classes $L_i$}
  \KwResult{Graph $\tilde{\mathcal{G}}$}
  \tcp{Subgraph Definition}
  \For{each image $I_i$}    
        {   
            $V_i =  O_i \cup C_i$ \\
            $E_i = \{ (u,v) \mid \forall u, v \in V_i\}$\\ 
        	$G_i=\{V_i, E_i\}$ \\
        }
        
  \tcp{Large Graph Definition}
    $L_i = \{l_{ij}\}_{j \in [1, ..., |O_i|]}$ \\
    $\tilde{O} = \cup_{G_i} O_i $\\ 
    $\tilde{V} = \cup_{G_i} V_i$ \\
    \For{all objects $o_s \in O_i$}    
        {   \For{all object $o_t \in O_m$} {
            $E_{st} = (v_s,v_t )\; \; \text{if} \, \,  l_{is} = l_{mt}$, \, \text{where} $v_s,v_t \in \tilde{V}$  \\}
        }
    $E_l = \{E_{st}\}_{s \in [1, ..., |\tilde{O}|], t \in [1, ..., |\tilde{O}|]}$ \\ 
    $\tilde{E} = (\cup_{G_i} E_i) \cup E_{l}$ \\
    $\tilde{\mathcal{G}} = \{ \tilde{V}, \tilde{E}\}$ \\
\caption{Graph Formulation for the MfM problem.}
\label{alg:algoritmo}
\end{algorithm}

To provide more insight in the process of encoding multiple local semantic maps into a graph structure, we highlight in Figure~\ref{fig:thegraph} the structure of the connectivity matrix, detailing how the objects are turned into nodes (\ref{fig:thegraph}.a); how to draw same-map (\ref{fig:thegraph}.b) and same-class (\ref{fig:thegraph}.c) connections; and how to initialize the embeddings (\ref{fig:thegraph}.d). The process is also summarized as pseudocode in Algorithm~\ref{alg:algoritmo}.

\section{Ablation on Loss Components}




In Section 3.3 of the main paper, we propose applying a combination of three loss functions (Euclidean Camera-Object Pose, Cross-Map Consistency, and Self-Similarity) to train the \emph{GNN-based Alignment Module}. In this section, we highlight the effect of 
varying the combinations of loss functions, as our goal is to explore the advantages of utilizing these three losses simultaneously. Using MfM with a TransformerGCN network, we show results  on both the MfM Dataset Small and Large scenes in the two opposite configurations: \textit{i)} graph with no known matches between the detections in different views and noisy local maps as input (Depth Local Maps + Class-based Correspondences), and \textit{ii)} graph with ground-truth detection matches and the ground truth local maps (GT Local Maps + GT Detection Matches).

\begin{table*}[t!]
\scriptsize
\centering
\begin{tabularx}{\textwidth}{X|ccc|ccc}
\hline
\multicolumn{7}{c}{Depth Local Maps + Class-based Correspondences} \\ \hline
\multirow{2}{*}{Method} & \multicolumn{3}{c|}{Small Scenes} & \multicolumn{3}{c}{Large Scenes} \\ \cline{2-7} 
 & Fail ($\%$) & $\mu_c \pm\sigma_c$ & $\mu_o \pm\sigma_o$ & Fail ($\%$) & $\mu_c \pm\sigma_c$ & $\mu_o \pm\sigma_o$ \\ \hline
 MfM Baseline  & 37 & 3.7 $\pm$ 2.2 & 3.5 $\pm$ 1.6 & 10  & 3.8 $\pm$ 1.8 & 2.7 $\pm$ 1.6 \\
MfM w/o Euclidean Cam-Obj Pose & 36 & 3.7 $\pm$ 2.1 & 3.5 $\pm$1.5 & 10 & 3.9  $\pm$ 1.7 & 2.6 $\pm$ 1.4 \\
MfM w/o Cross-Map Consistency & 37 & 3.5 $\pm$ 2.2 &  3.5 $\pm$1.5 & 10 & 3.7  $\pm$ 1.7 & 3.0 $\pm$ 1.6 \\ 
MfM w/o Self-Similarity & 40 & 3.8 $\pm$ 2.2 & 4.2 $\pm$1.5 & 10 & 3.8 $\pm$ 1.7 & 3.1 $\pm$ 1.5 \\ 
 \hline
 \hline
\multicolumn{7}{c}{GT Local Maps +  GT Detection Matches} \\ \hline
\multirow{2}{*}{Method} & \multicolumn{3}{c|}{Small Scenes} & \multicolumn{3}{c}{Large Scenes} \\ \cline{2-7} 
 & Fail ($\%$) & $ \mu_c(m)\pm\sigma_c(m)$ & $\mu_o(m) \pm\sigma_o(m)$ & Fail ($\%$) & $\mu_c(m)\pm\sigma_c(m)$ & $\mu_o(m)\pm\sigma_o(m)$ \\ \hline
 MfM Baseline & 31 & 3.9 $\pm$ 2.2 & 3.6 $\pm$ 1.4 & 5 & 3.7 $\pm$ 1.7 & 2.6 $\pm$ 1.2 \\
 MfM w/o Euclidean Cam-Obj Pose & 35 & 3.7 $\pm$ 2.2 & 3.5 $\pm$1.6 & 5 & 3.8 $\pm$ 1.8 & 2.4 $\pm$ 1.1  \\
 MfM w/o Cross-Map Consistency & 31 & 3.9 $\pm$ 2.2 & 3.6 $\pm$1.5 & 5 & 3.7 $\pm$ 1.9 & 3.1 $\pm$ 1.4 \\
 MfM w/o Self-Similarity & 31.9 & 3.7 $\pm$ 2.1 & 4.1 $\pm$ 1.4 & 5 & 4.2 $\pm$ 2.0 & 3.1 $\pm$ 1.7 \\
\end{tabularx}%
\caption{Given a set of local 2D semantic maps from
the Small and Large sequences of the MfM Dataset, we report the performance of MfM using \emph{i)} noisy inputs (Depth Local Maps + Class-based Correspondences) and \emph{ii)} perfect inputs (GT Local Maps +  GT Detection Matches).
Results include the average Euclidean error on the reconstructed object locations ($\mu_o$), and its standard deviation ($\sigma_0$). We report the fraction of scenes for which the reconstruction failed (Fail).}
\label{tab:ablation_real}
\end{table*}

As shown in Table~\ref{tab:ablation_real}, defining the loss function to incorporate all three losses proves to be advantageous for the overall predictive performance. Across both datasets and their various configurations, the MfM baseline, utilizing all three losses, consistently attains good results. Conversely, when the self-similarity loss is omitted, there is a noticeable decline in performance. The exclusion of either Euclidean Camera-Object Pose Loss or Cross-Map Consistency Loss leads to increased fluctuations in the results.


\section{Ablation on Synthetic Dataset}

In the main paper, we report results on synthetic scenes using $8$ cameras observing a scene with $7$ object of $5$ possible classes. These parameters were decided on empirical bases, after testing different combination of object and classes sizes. In this section, we summarize the results of such hyperparameter tuning.  
We set the visibility parameter to 1.0 ($\phi = 1.0$) and the noise level on the objects' locations in the local maps to 0 ($\Delta_{xy} = 0$). We report the results based on the average Euclidean error on the reconstructed object locations ($\mu_o$).

\begin{figure}[t!]
    \centering
        \includegraphics[width=\linewidth]{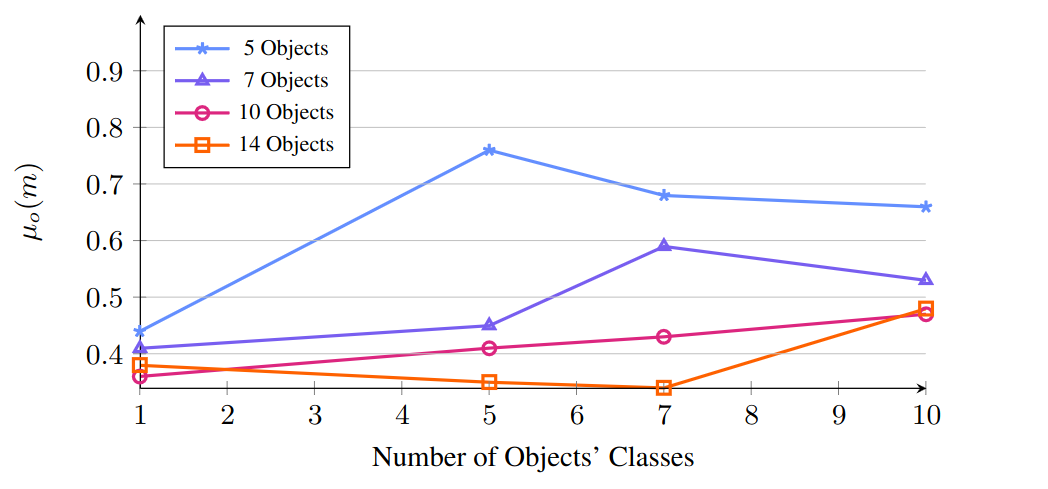}
    \caption{Average Euclidean error on the reconstructed object locations ($\mu_o$). 
    }
    \label{fig:ablation-synthetic} 
\end{figure}

The results, as illustrated in Figure~\ref{fig:ablation-synthetic}, reveal the trend where when the number of objects increases, the reconstructed object location error decreases. 
This observation underscores the correlation between the number of objects and the increased
accuracy in the process of object localization.

\section{Experiment Details}

\paragraph{Hardware.} The experiments were conducted on a machine with an NVIDIA RTX 4090 GPU, 64 GB RAM, and 12th Gen Intel(R) Core(TM) i9-12900KF CPU @ 3.20GHz.

\paragraph{Model Setting.} We train MfM in combination with different models using Adagrad as the optimization algorithm~\cite{duchi2011adaptive}. During our training process, we set a maximum of 1000 epochs, but we stopped the training earlier to prevent unnecessary iterations when the validation error no longer decreases. 
We optimized the weight decay using the \textit{Bayesian optimization} technique and obtained two different results based on the dataset:
\begin{itemize}
    \item \textbf{MfM Real Dataset.} We set the weight decay to 0.007.
    \item \textbf{MfM Synthetic Dataset.} We set the weight decay to 0.046.
\end{itemize}


\end{document}